\documentclass[review]{elsarticle}
\makeatletter
\def\ps@pprintTitle{%
 \let\@oddhead\@empty
 \let\@evenhead\@empty
 \def\@oddfoot{\centerline{\thepage}}%
 \let\@evenfoot\@oddfoot}
\makeatother

\usepackage[top=1.5in, bottom=1.5in, left=1.5in, right=1.5in]{geometry}
\usepackage{ragged2e}
\usepackage{graphicx} 
\graphicspath{{./}{./figures/}{./figures/final/}}
\usepackage{amsmath,epsfig}
\usepackage{algorithm}
\usepackage{algorithmicx}
\usepackage{amssymb} 
\usepackage{pifont} 
\usepackage{booktabs} 
\usepackage{multirow} 
\usepackage{caption}
\usepackage{color}
\usepackage[pagebackref=false,breaklinks=true, hidelinks,bookmarks=false]{hyperref}
\usepackage{epstopdf}
\usepackage{morefloats}
\usepackage{subcaption}
\DeclareGraphicsExtensions{.pdf,.eps,.png,.jpg,.mps}
\DeclareMathOperator\argmax{arg\,max}
\DeclareMathOperator\argmin{arg\,min}

\newcommand{\tabincell}[2]{\begin{tabular}{@{}#1@{}}#2\end{tabular}} 
\newcommand{\xmark}{\ding{53}} 
 \makeatletter
 \DeclareRobustCommand\onedot{\futurelet\@let@token\@onedot}
 \def\@onedot{\ifx\@let@token.\else.\null\fi\xspace}
\def\eg{\emph{e.g}\onedot} 
\def\ie{\emph{i.e}\onedot} 
 
 \def\etc{\emph{etc}\onedot} 
 
\def\etal{\emph{et al}\onedot}

\usepackage{xspace}

\usepackage{hyperref}

\begin{document}

\begin{frontmatter}

\title{Multiple Object Tracking: A Literature Review}






\author{Wenhan Luo$^{a,b}$, Junliang Xing$^c$, Anton Milan$^d$, Xiaoqin Zhang$^e$, Wei Liu$^a$, Tae-Kyun Kim$^b$}

\address{$^a$Tencent AI Lab, China \ \ \ \ $^b$Imperial College London, UK}
\address{$^c$Institute of Automation, Chinese Academy of Sciences, China}
\address{$^d$Amazon Research and Development Center, Germany \ \ $^e$Wenzhou University, China}

\begin{abstract}
Multiple Object Tracking (MOT) has gained increasing attention due to its academic and commercial potential. Although different approaches have been proposed to tackle this problem, it still remains challenging due to factors like abrupt appearance changes and severe object occlusions. In this work, we contribute the first comprehensive and most recent review on this problem. We inspect the recent advances in various aspects and propose some interesting directions for future research. To the best of our knowledge, there has not been any extensive review on this topic in the community. We endeavor to provide a thorough review on the development of this problem in recent decades. The main contributions of this review are fourfold: 1) Key aspects in an MOT system, including formulation, categorization, key principles, evaluation of MOT are discussed; 2) Instead of enumerating individual works, we discuss existing approaches according to various aspects, in each of which methods are divided into different groups and each group is discussed in detail for the principles, advances and drawbacks; 3) We examine experiments of existing publications and summarize results on popular datasets to provide quantitative and comprehensive comparisons. By analyzing the results from different perspectives, we have verified some basic agreements in the field; and 4) We provide a discussion about issues of MOT research, as well as some interesting directions which will become potential research effort in the future.
\end{abstract}


\end{frontmatter}


\section{Introduction}
\label{sec:introduction}
Multiple Object Tracking (MOT), or Multiple Target Tracking (MTT), plays an important role in computer vision. The task of MOT is largely partitioned into locating multiple objects, maintaining their identities, and yielding their individual trajectories given an input video. Objects to track can be, for example, pedestrians on the street \cite{yang2011learning,pellegrini2009you}, vehicles in the road \cite{koller1994robust,betke2000real}, sport players on the court \cite{lu2013learning, xing2011multiple, nillius2006multi}, or groups of animals (birds \cite{luo2014bilabel}, bats \cite{betke2007tracking}, ants \cite{khan2004mcmc}, fish \cite{spampinato2008detecting, spampinato2012covariance, fontaine2007model}, cells \cite{meijering2009tracking, li2008cell}, bees \cite{Bozek_2018_CVPR}, \etc). Multiple ``objects'' could also be viewed as different parts of a single object \cite{duan2012group}. In this review, we mainly focus on the research on pedestrian tracking. The underlying reasons for this specification are threefold. First, compared to other common objects in our environment, pedestrians are typical non-rigid objects, which is an ideal example to study the MOT problem. Second, videos of pedestrians arise in a huge number of practical applications, which further results in great commercial potential. Third, according to all data collected for this review, at least $70\%$ of current MOT research efforts are devoted to pedestrians.

As a mid-level task in computer vision, multiple object tracking grounds high-level tasks such as pose estimation \cite{ICCV15PoseVideo}, action recognition \cite{choi2012unified}, and behavior analysis \cite{hu2004survey}. It has numerous practical applications, such as visual surveillance \cite{wang2013intelligent}, human computer interaction \cite{candamo2010understanding} and virtual reality \cite{FCV12TrackAR}. These practical requirements have sparked enormous interest in this topic. Compared with Single Object Tracking (SOT), which primarily focuses on designing sophisticated appearance models and/or motion models to deal with challenging factors such as scale changes, out-of-plane rotations and illumination variations, multiple object tracking additionally requires two tasks to be solved: determining the number of objects, which typically varies over time, and maintaining their identities. Apart from the common challenges in both SOT and MOT, further key issues that complicate MOT include among others: 1) frequent occlusions, 2) initialization and termination of tracks, 3) similar appearance, and 4) interactions among multiple objects. In order to deal with all these issues, a wide range of solutions have been proposed in the past decades. These solutions concentrate on different aspects of an MOT system, making it difficult for MOT researchers, especially newcomers, to gain a comprehensive understanding of this problem. Therefore, in this work we provide a review to discuss the various aspects of the multiple object tracking problem.


\subsection{Differences from Other Related Reviews}
\label{difference}
To the best of our knowledge, there has not been any comprehensive literature review on the topic of multiple object tracking. However, there have been some other reviews related to multiple object tracking, which are listed in Table~\ref{table:other_reviews}. We group these surveys into four sets and highlight the differences from ours as follows.

\vspace{-0.4cm}
\begin{itemize}\setlength{\itemsep}{-0.2cm}
\item The first set \cite{zhan2008crowd,hu2004survey,kim2010intelligent,candamo2010understanding,wang2013intelligent} discusses tracking as an individual part while this work specifically discusses various aspects of MOT. For example, object tracking is discussed as a step in the procedure of high-level tasks such as crowd modeling \cite{zhan2008crowd,hu2004survey,kim2010intelligent}. Similarly, in \cite{candamo2010understanding} and \cite{wang2013intelligent}, object tracking is reviewed as a part of a system for behavior recognition \cite{candamo2010understanding} or video surveillance \cite{wang2013intelligent}.
\item The second set~\cite{forsyth2006computational,cannons2008review,yilmaz2006object,li2013survey} is dedicated to general visual tracking techniques~\cite{forsyth2006computational, cannons2008review, yilmaz2006object} or some special issues such as appearance models in visual tracking~\cite{li2013survey}. Their reviewing scope is wider than ours; ours on the contrary is more comprehensive and focused on multiple object tracking.
\item The third set~\cite{wu2013online,leal2015motchallenge} introduces and discusses benchmarks on general visual tracking~\cite{wu2013online} and on specific multiple object tracking~\cite{leal2015motchallenge}. Their attention is laid on experimental studies rather than literature reviews.
\item The fourth set \cite{zhao2019object} reviews the recent advances and development in object detection with the rising of deep learning. The topic is related to ours but different from ours. Object detection can provide observations for detection-based object tracking by locating the potential object locations in each frame, while MOT needs to associate these observations across multiple frames to form the object trajectories.
\end{itemize}
\vspace{-0.4cm}

\begin{table}[t]
\scriptsize
\caption{A summary of other literature reviews}
 \vspace{-0.35cm}
\label{table:other_reviews}
\tabcolsep=3pt
\centering
{\begin{tabular}{l l r}
\toprule
\textbf{Reference} & \textbf{Topic} & \textbf{Year}\\
\midrule
Zhan \etal~\cite{zhan2008crowd} & Crowd Analysis & 2008\\

Hu \etal~\cite{hu2004survey} & Object Motion and Behaviors & 2004\\

Kim \etal~\cite{kim2010intelligent} & Intelligent Visual Surveillance & 2010\\

Candamo \etal~\cite{candamo2010understanding} & Behavior Recognition in Transit Scenes & 2010\\

Xiaogang Wang~\cite{wang2013intelligent} & Multi-Camera Video Surveillance & 2013\\

\midrule
Forsyth \etal~\cite{forsyth2006computational} & Human Motion Analysis & 2006\\

Kevin Cannons~\cite{cannons2008review} & Visual Tracking & 2008\\

Yilmaz \etal~\cite{yilmaz2006object} & Object Visual Tracking & 2006\\

Li \etal~\cite{li2013survey} & Appearance Models in Object Tracking & 2013\\

\midrule
Wu \etal~\cite{wu2013online} & Visual Tracking Benchmark & 2013\\

Leal-Taix{\'e} \etal~\cite{leal2015motchallenge} & MOT Benchmark & 2015\\
\midrule
Zhao \etal~\cite{zhao2019object} & Object Detection & 2019\\
\bottomrule
\end{tabular}}
\end{table}

\subsection{Contributions}
\label{contribution}
We provide the first comprehensive review on the MOT problem to the computer vision community, which we believe is helpful to understand this problem, its main challenges, pitfalls, and the state of the art. The main contributions of this review are summarized as follows:
\vspace{-0.3cm}
\begin{itemize}\setlength{\itemsep}{-0.2cm}
	\item We derive a unified formulation of the MOT problem which consolidates most of the existing MOT methods (Section \ref{mot_formula}), and two different ways to categorize MOT methods (Section \ref{mot_categorization}).
	\item We investigate different key components involved in an MOT system, each of which is further divided into different aspects and discussed in detail regarding its principles, advances, and drawbacks (Section \ref{mot_components}).
	\item Experimental results on popular datasets regarding different approaches are presented, which makes future experimental comparison convenient. By investigating the provided results, some interesting observations and findings are revealed (Section \ref{mot_evaluation}).
	\item By summarizing the MOT review, we unveil existing issues of MOT research. Furthermore, open problems are discussed to identify potential future research directions (Section \ref{sec_summary}). 
\end{itemize}
\vspace{-0.3cm}
Note that this work is mainly dedicated to reviewing recent literature on the advances in multiple object tracking. As mentioned above, we also present experimental results on publicly available datasets excepted from existing publications to provide a quantitative view on the state-of-the-art MOT methods. For standardized benchmarking of multiple object tracking we kindly refer the readers to the recent work MOTChallenge by Leal-Taix\'e \etal \cite{leal2015motchallenge}.

\begin{table}[t]
\scriptsize
\caption{Denotations employed in this review}
 \vspace{-0.35cm}
\label{table:denotation}
\centering
\tabcolsep=3pt
{\begin{tabular}{c |c |c | c |c | c |c | c}
\hline
\textbf{Symbol} & \textbf{Description} & \textbf{Symbol} & \textbf{Description} & \textbf{Symbol} & \textbf{Description} & \textbf{Symbol} & \textbf{Description}\\
\hline
$P$ & probability & $\mathbf{I}$ & image & $\mathbf{p}$ & position & $x$, $y$ & position\\
$S$ & similarity & $\mathbf{S}$ & set of states & $\mathbf{v}$ & velocity & $u$, $v$ & speed \\
$C$ & cost & $\mathbf{O}$ & set of observations & $\mathbf{f}$ & feature & $w$, $\alpha$, $\lambda$ & weight \\
$N$ & frame number & $\mathbf{T}$ & trajectory/tracklet & $\mathbf{c}$ & color &t & time index \\
$M$ & object number& $\mathbf{M}$ & feature matrix & $\mathbf{o}$ & observation & $i$, $j$, $k$ & general index\\
$G$ & graph & $\mathbf{\Sigma}$ & covariance matrix & $\mathbf{s}$ & state & $\sigma$& variance \\
$V$ & vertex set & $\mathbf{L}$ & Laplacian matrix & $\mathbf{a}$ & acceleration &$\epsilon$ & noise \\
$E$ & edge set& $\mathbf{Y}$ & label set & $\mathbf{y}$ & label& $s$ & size\\
$D$ & distance & & & & & &\\
$L$ & likelihood & &  & & & &\\
$F$ & function & & & & & &\\
$Z$ & normalization factor & & & & & &\\
$\mathcal{N}$& normal distribution& & & & & &\\
$\mathcal{S}$& set & & & & & &\\
\hline

\end{tabular}}
\end{table}

\subsection{Organization of This Review}
\label{organization}
Our goal is to provide an overview of the major aspects in the MOT task. These aspects include the current state of research in MOT, all the detailed issues requiring consideration in building a system, and how to evaluate an MOT system. Section \ref{mot_problem} describes the MOT problem, including its general formulation (Section \ref{mot_formula}) and typical ways for categorization (Section \ref{mot_categorization}). Section \ref{mot_components} contributes to the most common components involved in modeling multi-object tracking, \ie, appearance model (Section \ref{ap_model}), motion model (Section \ref{mo_model}), interaction model (Section \ref{inter_model}), exclusion model (Section \ref{exc_model}), occlusion handling (Section \ref{occlusion}), and inference methods (Section \ref{inference}). Furthermore, issues concerning evaluations, including metrics (Section \ref{mot_metrics}), public datasets (Section \ref{datasets}), public codes (Section \ref{pub_algorithms}), and benchmark results (Section \ref{benchmark_results}) are discussed in Section \ref{mot_evaluation}. This part is followed by Section \ref{sec_summary} which summarizes the existing issues and interesting problems for future directions of the MOT research in the community.

\subsection{Denotations}
\label{mot_denotations}
Throughout this manuscript, we denote scalar and vector variables by lowercase letters (\eg,~$x$) and bold lowercase letters (\eg,~$\mathbf{x}$), respectively. We use bold capital letters (\eg,~$\mathbf{X}$) to denote a matrix or a set of vectors. Capital letters (\eg,~$X$) are adopted for specific functions or variables. Table \ref{table:denotation} lists symbols utilized throughout this review. Except the symbols in the table, there may be some symbols for a specific reference. As these symbols are not commonly employed, they are not listed in the table but will be rather defined in the context.

\section{MOT Problem}
\label{mot_problem}
We first endeavor to give a general mathematical formulation of MOT. We then discuss its possible categorizations based on different aspects.

\subsection{Problem Formulation}
\label{mot_formula}
The MOT problem has been formulated differently from various perspectives in previous works, which makes it difficult to understand this problem from a high-level view. Here we offer a general formulation and argue that existing works can be unified under this formulation. To the best of our knowledge, there has not been any previous work towards this attempt.

In general, multiple object tracking can be viewed as a multi-variable estimation problem. Given an image sequence, we employ $\mathbf{s}_t^i$ to denote the state of the $i$-th object in the $t$-th frame, $\mathbf{S}_t = (\mathbf{s}_t^1, \mathbf{s}_t^2, ..., \mathbf{s}_t^{M_t})$ to denote states of all the $M_t$ objects in the $t$-th frame. We employ $\mathbf{s}_{i_s:i_e}^i = \{\mathbf{s}_{i_s}^i, ..., \mathbf{s}_{i_e}^i\}$ to denote the sequential states of the \emph{i}-th object, where $i_s$ and $i_e$ are respectively the first and last frame in which target $i$ exists, and $\mathbf{S}_{1:t} = \{\mathbf{S}_1, \mathbf{S}_2, ..., \mathbf{S}_t\}$ to denote all the sequential states of all the objects from the first frame to the $t$-th frame. Note that the object number may vary from frame to frame.

Correspondingly, following the most commonly used tracking-by-detection, or Detection Based Tracking (DBT) paradigm, we utilize $\mathbf{o}_t^i$ to denote the collected observations for the $i$-th object in the $t$-th frame, $\mathbf{O}_t = (\mathbf{o}_t^1, \mathbf{o}_t^2, ..., \mathbf{o}_t^{M_t})$ to denote the collected observations for all the $M_t$ objects in the $t$-th frame, and $\mathbf{O}_{1:t} = \{\mathbf{O}_1, \mathbf{O}_2, ..., \mathbf{O}_t\}$ to denote all the collected sequential observations of all the objects from the first frame to the $t$-th frame.

The objective of multiple object tracking is to find the ``optimal'' sequential states of all the objects, which can be generally modeled by performing MAP (Maximum a posteriori) estimation from the conditional distribution of the sequential states given all the observations:
\begin{equation}
\label{eq:map}
\widehat{\mathbf{S}}_{1:t} = \underset{\mathbf{S}_{1:t}}\argmax \ P\left(\mathbf{S}_{1:t}|\mathbf{O}_{1:t}\right).
\end{equation}

Different MOT algorithms from previous works can now be thought as designing different approaches to solving the above MAP problem, either from a \emph{probabilistic inference} perspective \cite{xing2009multi,xing2011multiple,breitenstein2009robust,yang2009detection,mitzel2011real,rodriguez2011data,kratz2010tracking,reid1979algorithm} or a \emph{deterministic optimization} perspective \cite{zhang2008global, huang2008robust, li2009learning, pirsiavash2011globally, berclaz2011multiple, Butt13multi, shi2013multi, milan2014continuous, brendel2011multiobject, yang2012multi, andriyenko2012discrete, duan2012group}.

The probabilistic inference based approaches usually solve the MAP problem in Eqn. (\ref{eq:map}) using a two-step iterative procedure as follows,

\noindent\textbf{Predict}: $P(\mathbf{S}_t|\mathbf{O}_{1:t\!-\!1}\!)\!=\!\int\! P(\mathbf{S}_t|\mathbf{S}_{t\!-\!1})P(\mathbf{S}_{t\!-\!1}|\mathbf{O}_{1:t\!-\!1}) d\mathbf{S}_{\!t-\!1}$,

\noindent\textbf{Update}: $P(\mathbf{S}_t|\mathbf{O}_{1:t}) \propto P(\mathbf{O}_t|\mathbf{S}_t) P(\mathbf{S}_t|\mathbf{O}_{1:t-1})$.

\noindent{Here $P\left(\mathbf{S}_t|\mathbf{S}_{t-1}\right)$ and $P\left(\mathbf{O}_t|\mathbf{S}_t\right)$ are the \emph{Dynamic Model} and the \emph{Observation Model}, respectively.}

The deterministic optimization based approaches directly maximize the likelihood function $L\left(\mathbf{O}_{1:t}|\mathbf{S}_{1:t}\right)$ as a delegate of $P\left(\mathbf{S}_{1:t}|\mathbf{O}_{1:t}\right)$ over a set of available observations $\{\hat{\mathbf{O}}_{1:t}^n\}$:
\begin{equation}
\label{eq:ml}
\widehat{\mathbf{S}}_{1:t}= \underset{\mathbf{S}_{1:t}}\argmax \ P\left(\mathbf{S}_{1:t}|\mathbf{O}_{1:t}\right)= \underset{\mathbf{S}_{1:t}}\argmax \ L\left(\mathbf{O}_{1:t}|\mathbf{S}_{1:t}\right)= \underset{\mathbf{S}_{1:t}}\argmax\ {\prod_{n} P\left(\hat{\mathbf{O}}_{1:t}^n|\mathbf{S}_{1:t}\right)},
\end{equation}
or conversely minimize an energy function $E\left(\mathbf{S}_{1:t}|\mathbf{O}_{1:t}\right)$:
\begin{equation}
\label{eq:me}
\widehat{\mathbf{S}}_{1:t}= \underset{\mathbf{S}_{1:t}}\argmax \ P\left(\mathbf{S}_{1:t}|\mathbf{O}_{1:t}\right)= \underset{\mathbf{S}_{1:t}}\argmax \ \exp\left(-E\left(\mathbf{S}_{1:t}|\mathbf{O}_{1:t}\right)\right)/Z=\underset{\mathbf{S}_{1:t}}\argmin{\ E\left(\mathbf{S}_{1:t}|\mathbf{O}_{1:t}\right)},\\
\end{equation}
where $Z$ is a normalization factor to ensure $P\left(\mathbf{S}_{1:t}|\mathbf{O}_{1:t}\right)$ to be a probability distribution.

\begin{figure}[tb]
 \centering
 \includegraphics[width=0.85\linewidth]{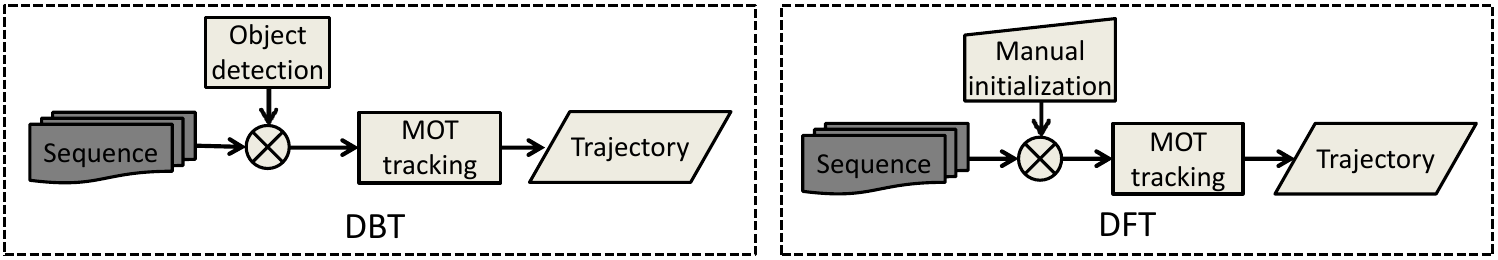}
 \vspace{-0.35cm}
 \caption{A procedure flow of two prominent tracking approaches. {\bf Left:} Detection-Based Tracking (DBT), {\bf right:} Detection-Free Tracking (DFT).}
 \label{fig:dbt_dft}
\end{figure}

\subsection{MOT Categorization}
\label{mot_categorization}
It is difficult to classify one particular MOT method into a distinct category by a universal criterion. Admitting this, it is thus feasible to group MOT methods by multiple criteria. In the following we attempt to conduct this according to three criteria: \emph{a) initialization method}, \emph{b) processing mode}, and \emph{c) type of output}. 
The reason we choose these three criteria is that this naturally follows the way of processing a task, \ie, how the task is initialized, how it is processed and what type of result is obtained. We believe that other criteria may also be reasonably adopted to categorize various MOT methods. However, categorizing different MOT methods with all possible criteria are beyond the scope of this article. In the following, each of the above criteria along with its corresponding categorization is represented.

\subsubsection{Initialization Method}
\label{dbt_vs_dft}
Most existing MOT works can be grouped into two sets~\cite{yang2012onlinelearned}, depending on how objects are initialized: Detection-Based Tracking (DBT) and Detection-Free Tracking (DFT).

\emph{\textbf{Detection-Based Tracking}}. As shown in Figure~\ref{fig:dbt_dft} (top), objects are first detected and then linked into trajectories. This strategy is also commonly referred to as ``tracking-by-detection''. Given a sequence, type-specific object detection or motion detection (based on background modeling)~\cite{bose2007multi,song2010stochastic} is applied in each frame to obtain object hypotheses, then (sequential or batch) tracking is conducted to link detection hypotheses into trajectories. There are two issues worth noting. First, since the object detector is trained in advance, the majority of DBT focuses on specific kinds of targets, such as pedestrians, vehicles or faces. Second, the performance of DBT highly depends on the performance of the employed object detector.

\emph{\textbf{Detection-Free Tracking}}. As shown in Figure~\ref{fig:dbt_dft} (bottom), DFT~\cite{hu2012single,zhang2013structure,zhang2013preserving,yang2007game} requires manual initialization of a fixed number of objects in the first frame, then localizes these objects in subsequent frames.

DBT is more popular because new objects are discovered and disappearing objects are terminated automatically. DFT cannot deal with the case that objects appear. However, it is free of pre-trained object detectors. Table \ref{table:dbt_dft} lists the major differences between DBT and DFT.

\begin{table}[t]
\scriptsize
\caption{Comparison between DBT and DFT. Adapted from~\cite{yang2012onlinelearned}.}
 \vspace{-0.35cm}
\label{table:dbt_dft}
\centering
{\begin{tabular}{c c c}
\hline
\textbf{Item} & \textbf{DBT} & \textbf{DFT} \\
\hline
Initialization & automatic, imperfect & manual, perfect \\
\hline
\# of objects & varying & fixed \\
\hline
Applications & \tabincell{c}{specific type of objects (in most cases)} & any type of objects \\
\hline
Advantages & \tabincell{c}{ability to handle varying number of objects} & free of object detector \\
\hline
Drawbacks & \tabincell{c}{performance depends on object detection} & manual initialization \\
\hline
\end{tabular}}
\end{table}

\begin{figure}[t]
 \centering
 \includegraphics[width=0.9\linewidth]{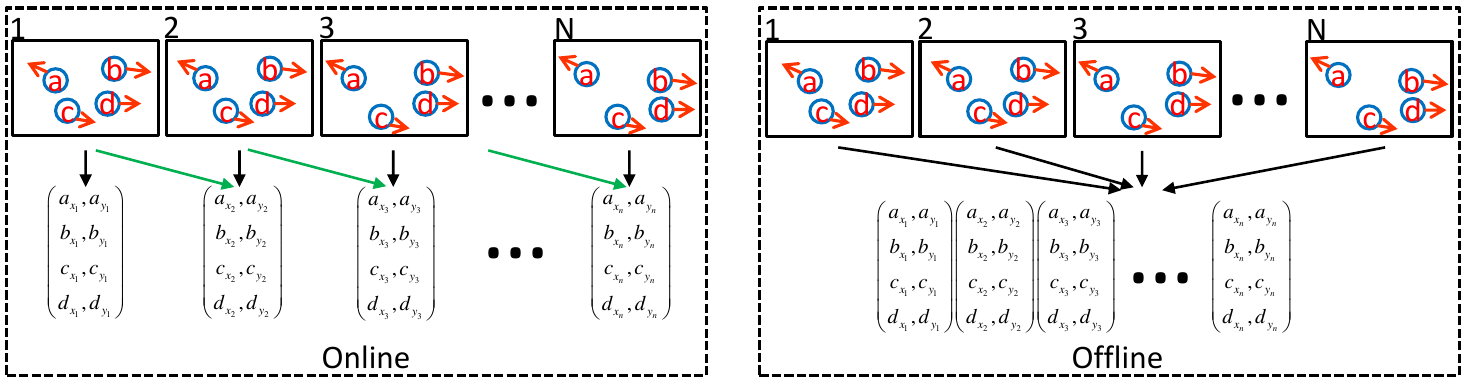}
\vspace{-0.35cm}
 \caption{An illustration of online (left) and offline (right) tracking. Best viewed in color.}
 \label{fig:seq_bat}
\end{figure}

\subsubsection{Processing Mode}
\label{online_vs_offline}
MOT can also be categorized into \emph{online} tracking and \emph{offline} tracking. The difference is whether observations from future frames are utilized when handling the current frame. Online, also called causal, tracking methods only rely on the past information available up to the current frame, while offline, or batch tracking approaches employ observations both in the past and in the future.

\emph{\textbf{Online tracking}}. In online tracking \cite{hu2012single, zhang2012online, zhang2013structure, zhang2013preserving, xiang2015learning,Yoon_2016_CVPR}, the image sequence is handled in a step-wise manner, thus online tracking is also named as sequential tracking. An illustration is shown in Figure \ref{fig:seq_bat} (top), with three objects (different circles) a, b, and c. The green arrows represent observations in the past. The results are represented by the object's location and its ID. Based on the up-to-time observations, trajectories are produced on the fly.

\emph{\textbf{Offline tracking}}. Offline tracking \cite{song2010stochastic, qin2012improving, yang2012multi, yang2012online, brendel2011multiobject, yang2011learning, kuo2010multi, henriques2011globally, sugimura2009using, choi2010multiple} utilizes a batch of frames to process the data. As shown in Figure \ref{fig:seq_bat} (bottom), observations from all the frames are required to be obtained in advance and are analyzed jointly to estimate the final output. Note that, due to computational and memory limitation, it is not always possible to handle all the frames at once. An alternative solution is to split the data into shorter video clips, and infer the results hierarchically or sequentially for each batch. Table \ref{table:seq_bat} lists the differences between the two processing modes.

\subsubsection{Type of Output}
\label{type_output}

This criterion classifies MOT methods into deterministic ones and probabilistic ones, depending on the randomness of output. The difference between these two types of methods primarily results from the optimization methods adopted as mentioned in Section \ref{mot_formula}. 

\emph{\textbf{Stochastic Tracking}}. The output results of stochastic tracking vary from time to time. For example, in the case of detection-free tracking, the bounding box results are different if we utilize particle filter for inference. The difference results from the randomness of the generation of particles in the processing. Even in the case of detection-based tracking, some studies also employ state-of-the-art single object tracker to refine the detection bounding boxes. This kind of methods will also lead to different tracking results in different running times.

\emph{\textbf{Deterministic Tracking}}. The output of deterministic tracking is constant when running the methods multiple times. For instance, in the case of tracking-by-detection, data association methods like Hungarian algorithm will produce deterministic tracking results. Deterministic tracking usually is associated with deterministic optimization for deriving the final output.

\subsubsection{Discussion}
The difference between DBT and DFT is whether a detection model is adopted (DBT) or not (DFT). The key to differentiate online and offline tracking is the way they process observations. Readers may question whether DFT is identical to online tracking because it seems DFT always processes observations sequentially. This is true in most cases although some exceptions exist. Orderless tracking \cite{hong2013orderless} is an example. It is DFT and simultaneously processes observations in an orderless way. Though it is for single object tracking, it can also be applied for MOT, and thus DFT can also be applied in a batch mode. Another vagueness may rise between DBT and offline tracking, as in DBT tracklets or detection responses are usually associated in a batch way. Note that there are also sequential DBT which conducts association between previously obtained trajectories and new detection responses \cite{luo2014bilabel,luo2013generic,xing2009multi}.

The categories presented above in Section \ref{dbt_vs_dft}, \ref{online_vs_offline} and \ref{type_output} are three possible ways to classify MOT methods, while there may be others. Notably, specific solutions for sport scenarios \cite{xing2011multiple,lu2013learning}, aerial scenes \cite{reilly2010detection,shi2013multi}, generic objects \cite{luo2013generic,dicle2013way,brostow2006unsupervised,luo2014bilabel,Sekii_2016_CVPR}, \etc exist\textcolor{blue}{,} and we suggest the readers refer to the respective publications.

By providing these three criteria described above, it is convenient for one to tag a specific method with the combination of the categorization label. This would help one to understand a specific approach easier.

\section{MOT Components}
\label{mot_components}
In this section, we represent the primary components of an MOT approach. As mentioned above, the goal of MOT is to discover multiple objects in individual frames and recover the identity information across continuous frames, \ie, trajectory, from a given sequence. When developing MOT approaches, two major issues should be considered. One is how to measure similarity between objects in frames, the other one is how to recover the identity information based on the similarity measurement between objects across frames. Roughly speaking, the first issue involves the modeling of appearance, motion, interaction, exclusion, and occlusion. The second one involves with the inference problem. We review recent progress regarding both items in the following.

\begin{table}[t]
\scriptsize
\caption{Comparison between online and offline tracking\label{table:seq_bat}}
 \vspace{-0.35cm}
\centering
{\begin{tabular}{ccc}
\toprule
\textbf{Item} & \textbf{Online tracking} & \textbf{Offline tracking}\\
\midrule
Input& Up-to-time observations & All observations \\
\midrule
Methodology & \tabincell{c}{Gradually extend existing trajectories\\ with current observations} & \tabincell{c}{Link observations into trajectories}\\
\midrule
Advantages & \tabincell{c}{Suitable for online tasks} & \tabincell{c}{Obtain global optimal solution theoretically}\\
\midrule
Drawbacks& \tabincell{c}{Suffer from shortage of observation}& \tabincell{c}{Delay in outputting final results} \\
\bottomrule
\end{tabular}}
\end{table}

\subsection{Appearance Model}
\label{ap_model}
Appearance is an important cue for affinity computation in MOT. However, different from single object tracking, which primarily focuses on constructing a sophisticated appearance model to discriminate object from background, most MOT methods do not consider appearance modeling as the core component, although it can be an important one.

Technically, an appearance model includes two components: \emph{visual representation} and \emph{statistical measuring}. Visual representation describes the visual characteristics of an object using some features, either based on a single cue or multiple cues. Statistical measuring, on the other hand, is the computation of similarity between different observations. More formally, the similarity between two observations $i$ and $j$ can be written as
\begin{equation}
\label{eq:ap_model}
S_{ij} = F\left(\mathbf{o}_i,\mathbf{o}_j\right),
\end{equation}
where $\mathbf{o}_i$ and $\mathbf{o}_j$ are visual representations of different observations, and $F\left(\cdot,\cdot\right)$ is a function that measures the similarity between them. In the following, we first discuss the visual representation in MOT, and then describe statistical measurement, respectively.

\begin{figure}[tb]
 \centering
 \includegraphics[width=0.85\linewidth]{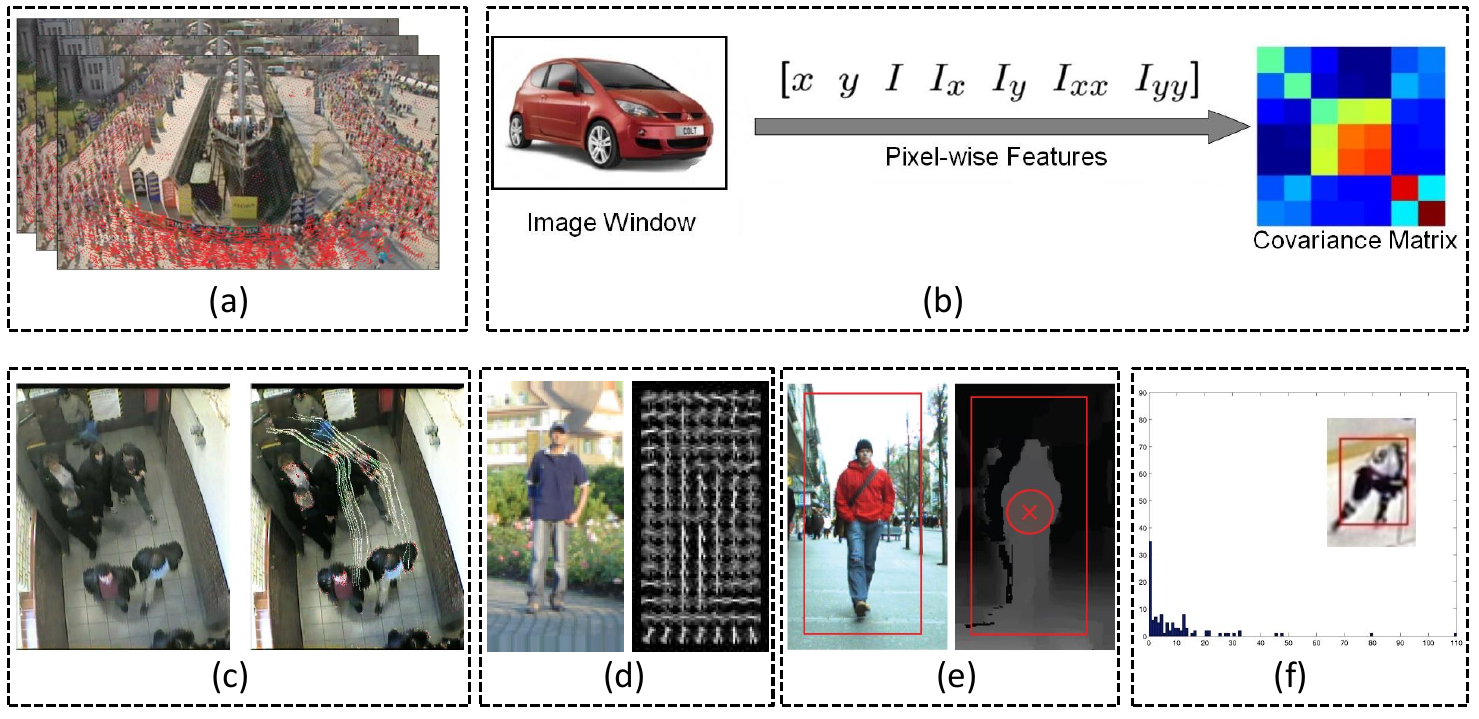}
\vspace{-0.35cm}
 \caption{An illustration of various visual features. (a) Optical flow \cite{ali2008floor}, (b) covariance matrix, (c) point features \cite{brostow2006unsupervised}, (d) gradient based features (HOG) \cite{dalal2005histograms}, (e) depth \cite{mitzel2010multi}, and (f) color features \cite{okuma2004boosted}. Best viewed in color.} 
\label{fig:feature_examples}
\end{figure}

\subsubsection{Visual Representation}
\label{features}
Visual representation describes an object according to different kinds of features, which are shown in Figure~\ref{fig:feature_examples}. We group features into the following different categories.

\textbf{Local features.} KLT is an example of searching ``good'' local features and tracking. It is successfully adopted in both SOT \cite{shi1994good} and MOT. Obtaining easy-to-track features, we can employ them to generate short trajectories \cite{sugimura2009using,zhao2012tracking}, estimate camera motion \cite{choi2010multiple,benfold2011stable}, motion clustering \cite{brostow2006unsupervised} and so on. Optical flow can also be regarded as local features if we treat image pixel as the finest local range. A set of solutions to MOT utilize optical flow to link detection responses into short tracklets before data association \cite{rodriguez2009tracking,izadinia2012}. As it is related with motion, it is utilized to encode motion information \cite{walk2010new, choi2015near}. One special application of optical flow is to discover crowd motion patterns in packed scenarios \cite{ali2008floor,rodriguez2011data}, where ordinary features are not reliable.

\textbf{Region features.} Compared with local features, region features are extracted from a wider range (\eg a bounding box). We illustrate them as three types: a) \emph{zero-order} type, b) \emph{first-order} type and c) \emph{up-to-second-order} type. Here, order means the order of discrepancy when computing the representation. For instance, zero-order means values of pixels are not compared, while one-order means discrepancy values among pixels are computed once.
\vspace{-0.3cm}
\begin{itemize}\setlength{\itemsep}{-0.2cm}
	\item \emph{Zero-order.} This is the most widely utilized representation for MOT. Color histogram \cite{sugimura2009using,mitzel2010multi,izadinia2012,okuma2004boosted,mitzel2011real,Yu_2016_CVPR} and raw pixel template \cite{yamaguchi2011you} are two typical examples of this type. 
	\item \emph{First-order.} Gradient-based representations like HOG \cite{izadinia2012,kuo2010multi, breitenstein2009robust, choi2012unified, yu2008distributed} and level-set formulation \cite{mitzel2010multi} are commonly employed.
	\item \emph{Up-to-second-order.} Region covariance matrix~\cite{porikli2006covariance,tuzel2006region} which computes up-to-second-order discrepancy belongs to this set. It has been adopted in \cite{henriques2011globally, kuo2010multi,hu2012single}.
\end{itemize}
\vspace{-0.3cm}
\textbf{Others.} Besides local and region features, there are some other kinds of representation. Taking depth as an example, it is typically used to refine detection hypotheses \cite{mitzel2010multi, ess2009robust, ess2007depth, ess2008mobile, gavrila2007multi}. The Probabilistic Occupancy Map (POM) \cite{fleuret2008multicamera,berclaz2011multiple} is employed to estimate how likely an object would occur in a specific grid cell. One more example is gait feature, which is unique for individual persons \cite{sugimura2009using}. DCNN \cite{xu2014person} plays a role of a codebook similar to bag-of-words (BoW) in \cite{Xu_2016_CVPR}. ColorNames descriptor is utilized in \cite{Xu_2018_CVPR} for appearance representation. Deep features from Convolution Neural Network (CNN) are employed for visual representation in \cite{Kim_2018_ECCV,He_2019_CVPR}. In \cite{Zhang_2019_ICCV}, point cloud feature is the first time to be introduced for feature fusion in MOT.

\textbf{Discussion.} Generally, color histogram is a well studied similarity measure, but it ignores the spatial layout of the object region. Local features are efficient, but sensitive to issues like occlusion and out-of-plane rotation. Gradient based features like HOG can describe the shape of an object and are robust to certain transformations such as illumination changes, but they cannot handle occlusion and deformation well. Region covariance matrix features are more robust as they take more information in account, but this benefit is obtained at the cost of more computation. Depth features make the computation of affinity more accurate, but they require multiple views of the same scenery and/or additional algorithm \cite{felzenszwalb2006efficient} to obtain depth measurements.


\subsubsection{Statistical Measuring}
This step is closely related to the section above. Based on visual representation, statistical measure computes the affinity between two observations. While some approaches solely rely on one kind of cue, others are built on multiple cues.

\textbf{Single cue.} Modeling appearance using single cue is either transforming distance into similarity or directly calculating the affinity. For example, the Normalized Cross Correlation (NCC) is usually adopted to calculate the affinity between two counterparts based on the representation of raw pixel template mentioned above \cite{yamaguchi2011you, ali2008floor, wu2012coupling, pellegrini2009you}. Speaking of color histogram, Bhattacharyya distance $B\left(\cdot,\cdot\right)$ is used to compute the distance between two color histograms $\mathbf{c}_i$ and $\mathbf{c}_j$. The distance is transformed into similarity $S$ like $S\left(\mathbf{T}_i,\mathbf{T}_j\right) = \exp\left(-B\left(\mathbf{c}_i,\mathbf{c}_j\right)\right)$ \cite{kratz2010tracking, sugimura2009using, choi2010multiple, leibe2008coupled, qin2012improving, xing2009multi} or fit the distance to Gaussian distributions like \cite{zhang2008global}. Transformation of dissimilarity into likelihood is also applied to the representation of covariance matrix \cite{henriques2011globally}. Cosine similarity between deep features from neural network is used in \cite{Ren_2018_ECCV}. Besides these typical models, bag-of-words model \cite{lazebnik2006beyond} is employed based on point feature representation \cite{yang2009detection}.

\textbf{Multiple cues.} Different kinds of cues can complement each other to make the appearance model more robust. However, it not trivial to decide how to fuse the information from multiple cues. Regarding this, we summarize multi-cue based appearance models according to five kinds of fusion strategies: \textit{Boosting}, \textit{Concatenating}, \textit{Summation}, \textit{Product}, and \textit{Cascading} (see also Table~\ref{table:multi_cue_overview}).
\vspace{-0.3cm}
\begin{itemize}\setlength{\itemsep}{-0.2cm}
	\item Boosting. The strategy of Boosting usually selects a portion of features from a feature pool sequentially via a Boosting based algorithm. For example, from color histogram, HOG and covariance matrix descriptor, AdaBoost, RealBoost, and a HybridBoost algorithm are respectively employed to choose the most representative features to discriminate pairs of tracklets of the same object from those of different objects in \cite{kuo2010multi}, \cite{yang2012onlinelearned} and \cite{li2009learning}. 
	
	\item Concatenation. Different kinds of features can be concatenated for computation. In \cite{brendel2011multiobject}, color, HOG and optical flow are concatenated for appearance modeling.
	
	\item Summation. This strategy takes affinity values from different features and balances these values with weights \cite{mitzel2010multi,liu2012automatic,takala2007multi}. 
	
	\item Product. Differently from the strategy above, values are multiplied to produce the integrated affinity \cite{yang2009detection,song2010stochastic,giebel2004bayesian,berclaz2006robust}. Note that, independence assumption is usually made when applying this strategy.
	
	\item Cascading. This is a cascade manner of using various types of visual representation, either to narrow the search space \cite{gavrila2007multi} or model appearance in a coarse-to-fine way \cite{izadinia2012}.
\end{itemize}
\vspace{-0.3cm}

\begin{table}[t]
\scriptsize
\caption{An overview of typical appearance models employing multiple cues.}
 \vspace{-0.35cm}
\label{table:multi_cue_overview}
\centering
{\begin{tabular}{c| c | c}
\hline
\textbf{Strategy} & \textbf{Employed Cue} & \textbf{Ref.} \\
\hline
Boosting & \tabincell{c}{Color, HOG, shapes, covariance matrix, \etc} & \cite{kuo2010multi,li2009learning,yang2012onlinelearned}\\
\hline
Concatenating &\tabincell{c}{Color, HOG, optical flow, \etc} & \cite{brendel2011multiobject}\\
\hline
Summation & \tabincell{c}{Color, depth, correlogram, LBP, \etc} & \cite{mitzel2010multi,liu2012automatic,takala2007multi}\\
\hline
Product & \tabincell{c}{Color, shapes, bags of local features, \etc} & \cite{yang2009detection,song2010stochastic,giebel2004bayesian,berclaz2006robust}\\
\hline
Cascading & \tabincell{c}{Depth, shape, texture, \etc} & \cite{gavrila2007multi,izadinia2012}\\
\hline
\end{tabular}}
\end{table}

\subsection{Motion Model}
\label{mo_model}
The motion model captures the dynamic behavior of an object. It estimates the potential position of objects in the future frames, thereby reducing the search space. In most cases, objects are assumed to move smoothly in the world and therefore in the image space (except for abrupt motions). We will discuss linear motion model and non-linear motion model in the following.

\begin{figure}[t]
 \centering
 \includegraphics[width=\linewidth]{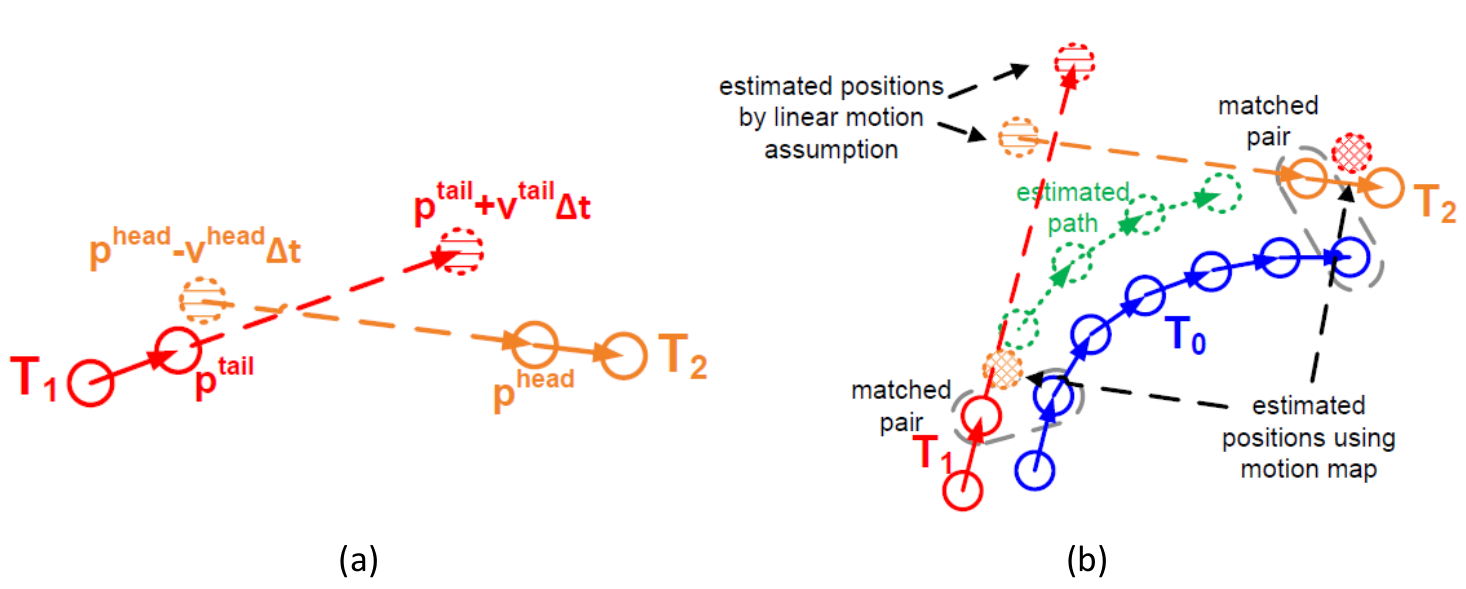}
\vspace{-0.35cm}
 \caption{An image comparing the linear motion model (a) with the non-linear motion model (b) \cite{yang2012multi}. Best viewed in color.}
 \label{fig:motion_ex2}
\end{figure}

\subsubsection{Linear Motion Model}
\label{li_mo_model}
This is by far the most popular model~\cite{shafique2008rank,yu2007multiple,breitenstein2009robust}. A constant velocity assumption \cite{breitenstein2009robust} is made in this model. Based on this assumption, there are three different ways to construct the model.
\vspace{-0.3cm}
\begin{itemize}\setlength{\itemsep}{-0.2cm}
\item Velocity smoothness is modeled by enforcing the velocity values of an object in successive frames to change smoothly. In \cite{milan2014continuous}, it is implemented as a cost term,
\begin{equation}
\label{eq:mo2}
C_{dyn}= \sum_{t=1}^{N-2}\sum_{i=1}^M\left\|\mathbf{v}_i^t-\mathbf{v}_i^{t+1}\right\|^2,
\end{equation}
where the summation is conducted over $N$ frames and $M$ trajectories/objects.

\item Position smoothness directly forces the discrepancy between the observed position and estimated position. Let us take \cite{xing2009multi} as an example. Considering a temporal gap $\Delta t$ between tail of tracklet $\mathbf{T}_i$ and head of tracklet $\mathbf{T}_j$, the smoothness is modeled by fitting the estimated position to a Gaussian distribution with the observed position as center. In the stage of estimation, both forward motion and backward motion are considered. Thus, the affinity considering linear motion model is,

\begin{equation}
\label{eq:mo3}
P_m\left(\mathbf{T}_i,\mathbf{T}_j\right) = \mathcal{N}\left(\mathbf{p}_i^{tail} + \mathbf{v}_i^F\Delta t;\mathbf{p}_j^{head},\Sigma_j^B\right)\ast \mathcal{N}\left(\mathbf{p}_j^{head} + \mathbf{v}_i^B\Delta t;\mathbf{p}_i^{tail},\Sigma_i^F\right).\\
\end{equation}
where ``F'' and ``B'' means forward and backward direction. A similar strategy is adopted by Yang \etal \cite{yang2012online}. The displacement between observed position and estimated position $\Delta \textbf{p}$ is fit to a Gaussian distribution with zero center. Other examples of this strategy are \cite{kuo2010multi,kuo2011does,yang2011learning,qin2012improving,nillius2006multi,yang2012online}. 
	
	\item Acceleration smoothness. Besides considering position and velocity smoothness, acceleration is taken into account \cite{kuo2011does}. The probability distribution of motion of a state $\{\hat{\mathbf{s}}_k\}$ at time $k$ given the observation tracklet $\{\mathbf{o}_k\}$ is modeled as,
\begin{equation}
\label{eq:mo5}
P\left(\left\{\hat{\mathbf{s}}_k\right\}|\left\{\mathbf{o}_k\right\}\right) = \prod_k \mathcal{N}\left(\mathbf{x}_k-\hat{\mathbf{x}}_k; \mathbf{0}, \Sigma_p\right)\prod_k \mathcal{N}\left(\mathbf{v}_k; \mathbf{0}, \Sigma_\mathbf{v}\right) \prod_k \mathcal{N}\left(\mathbf{a}_k;\mathbf{0},\Sigma_\mathbf{a}\right),\\
\end{equation}
where $\mathbf{v}_k$ is the velocity, $\mathbf{a}_k$ is the acceleration, and $\mathcal{N}$ is a zero-mean Gaussian distribution.
\end{itemize}
\vspace{-0.3cm}

\subsubsection{Non-linear Motion Model}
\label{noli_mo_model}
The linear motion model is commonly used to explain the object's dynamics. However, there are some cases which the linear motion model cannot deal with. To this end, non-linear motion models are proposed to produce more accurate motion affinity between tracklets. For instance, Yang \etal \cite{yang2012multi} employ a non-linear motion model to handle the situation that targets may move freely. Given two tracklets $\mathbf{T}_1$ and $\mathbf{T}_2$ which belong to the same target in Figure~\ref{fig:motion_ex2}(a), the linear motion model~\cite{yang2012online} would produce a low probability to link them. Alternatively, employing the non-linear motion model, the gap between the tail of tracklet $\mathbf{T}_1$ and the head of tracklet $\mathbf{T}_2$ could be reasonably explained by a tracklet $\mathbf{T}_0 \in \mathcal{S}$, where $\mathcal{S}$ is the set of support tracklets. As shown in Figure~\ref{fig:motion_ex2}(b), $\mathbf{T}_0$ matches the tail of $\mathbf{T}_1$ and the head of $\mathbf{T}_2$. Then the real path to bridge $\mathbf{T}_1$ and $\mathbf{T}_2$ is estimated based on $\mathbf{T}_0$, and the affinity between $\mathbf{T}_1$ and $\mathbf{T}_2$ is computed similarly as described in Section~\ref{li_mo_model}.

\subsection{Interaction Model}
\label{inter_model}
Interaction model, also known as mutual motion model, captures the influence of an object on other objects. In the crowd scenery \cite{Yu_2016_CVPR}, an object would experience some ``force'' from other agents and objects. For instance, when a pedestrian is walking on the street, he would adjust his speed, direction and destination, in order to avoid collisions with others. Another example is when a crowd of people walk across a street, each of them follows other people and guides others at the same time. In fact, these are examples of two typical interaction models known as the \emph{social force models} \cite{helbing1995social} and the \emph{crowd motion pattern models} \cite{hu2008detecting}.

\subsubsection{Social Force Models}
\label{social_model}
Social force models are also known as group models. In these models, each object is considered to be dependent on other objects and environmental factors. This type of information could alleviate performance deterioration in crowded scenes. In social force models, targets are considered as agents which determine their velocity, acceleration, and destination based on observations of other objects and the environment. More specifically, in social force models, target behavior is modeled based on two aspects, \emph{individual force} and \emph{group force}.

\textbf{Individual force.} For each individual in a group of multiple objects, two types of forces are considered:
\vspace{-0.3cm}
\begin{itemize}\setlength{\itemsep}{-0.2cm}
	\item \emph{\textbf{fidelity}}, which means one should not change his desired destination
	\item \emph{\textbf{constancy}}, which means one should not suddenly change his momentum, including speed and direction
\end{itemize}
\vspace{-0.3cm}

\textbf{Group force.} For a whole group, three types of forces are considered: 
\vspace{-0.3cm}
\begin{itemize}\setlength{\itemsep}{-0.2cm}
	\item \emph{\textbf{attraction}}, which means individuals moving together as a group should stay close
	\item \emph{\textbf{repulsion}}, which means that individuals moving together as a group should keep some distance away from others to make all members comfortable
	\item \emph{\textbf{coherence}}, which means individuals moving together as a group should move with similar velocity
\end{itemize}
\vspace{-0.3cm}
The majority of existing publications modeling interaction among objects with social force typically minimizes an energy objective, consisting of terms reflecting individual force and group force. Table \ref{table:social_force_examples} lists exemplar publications in the community which adopt social force models for interaction modeling. While \cite{Maksai_2019_CVPR} is an exception of explicitly modeling social force as energy terms. The social force is encoded as the so-called social feature for further processing in this study.

\begin{table}[t]
\scriptsize
\caption{Existing publications employing social force models.}
 \vspace{-0.35cm}
\label{table:social_force_examples}
\tabcolsep=15pt
\centering
{\begin{tabular}{c|c}
\hline
 \textbf{Ref.} & \textbf{Employed Forces}\\
\hline
\cite{pellegrini2009you} & \tabincell{c}{repulsion, constancy, fidelity}\\
\hline
\cite{yamaguchi2011you}  & \tabincell{c}{repulsion, constancy, attraction, coherence}\\
\hline
\cite{qin2012improving} & \tabincell{c}{coherence}\\
\hline
\cite{choi2010multiple}  & \tabincell{c}{repulsion, coherence}\\
\hline
\cite{scovanner2009learning}  & \tabincell{c}{constancy, fidelity, repulsion}\\
\hline
 \cite{pellegrini2010improving} & \tabincell{c}{constancy, coherence}\\
\hline
\end{tabular}}
\end{table}

\subsubsection{Crowd Motion Pattern Models}
\label{crowd_model}
Inspired by the crowd simulation literature \cite{zhan2008crowd}, motion patterns are introduced to alleviate the difficulty of tracking an individual object in the crowd. In general, this type of models is usually applied in the over-crowded scenario where the density of targets is considerably high. In such highly-crowded scenery, objects are usually quite small, and cues such as appearance and individual motion are ambiguous. In this case, motion from the crowd is a comparably reliable cue for the problem.

Roughly, there are two kinds of motion patterns, \emph{structured} and \emph{unstructured} ones. Structured motion patterns exhibit collective spatio-temporal structure while unstructured motion patterns exhibit various modalities of motion. In general, motion patterns are learned by various methods (including ND tensor voting \cite{zhao2012tracking}, Hidden Markov Models \cite{kratz2010tracking,kratz2012tracking}, Correlated Topic Model \cite{rodriguez2009tracking}, sometimes considering scene structures \cite{ali2008floor}) and applied as prior knowledge to assist object tracking.

\subsection{Exclusion Model}
\label{exc_model}
Exclusion is a constraint employed to avoid physical collisions when seeking a solution to the MOT problem. It arises from the fact that two distinct objects cannot occupy the same physical space in the real world. Given multiple detection responses and multiple trajectory hypotheses, generally there are two constraints. The first one is the so-called \emph{detection-level exclusion} \cite{milan13detection}, \ie, two different detection responses in the same frame cannot be assigned to the same target. The second one is the so-called \emph{trajectory-level exclusion}, \ie, two trajectories cannot be infinitely close to each other.

\subsubsection{Detection-level Exclusion Modeling}
Different approaches are adopted to model the detection-level exclusion. Basically, there are ``soft'' and ``hard'' models. 

\textbf{``Soft'' modeling.} Detection-level exclusion is ``softly'' modeled by minimizing a cost term to penalize the case of violation. For example, a penalty is defined if two simultaneous detection responses are assigned the same label of trajectory and they are sufficiently distant from each other in \cite{milan13detection}.

To model exclusion, a special exclusion graph is constructed to capture the constraint \cite{kc2013discriminative}. Given all the detection responses, they define a graph where nodes represent detection responses. Each node (one detection) is connected only to nodes (other detections) that exist at the same time as the node itself. After constructing this graph, the label assignment is maximized \emph{w.r.t.} exclusion to encourage connected nodes to have different labels as $\operatorname{Tr}\left(\mathbf{Y}\mathbf{L}\mathbf{Y}\right)$, where $\mathbf{L}$ is the Laplacian matrix, $\mathbf{Y} = \left(\mathbf{y}_1,...,\mathbf{y}_{|V|}\right)$ is the label assignment of all the $|V|$ nodes in the graph, and $\operatorname{Tr}\left(\cdot\right)$ is the trace norm of a matrix.

\textbf{``Hard'' modeling.} ``Hard'' modeling of detection-level exclusion is implemented by applying explicit constraint. For instance, to model the detection-level exclusion, the so-called cannot links are introduced to imitate that if two tracklets have overlap in their time span, then they cannot be assigned to the same cluster, \ie to belong to the same trajectory \cite{luo2015automatic}. Non-negative discretization is conducted in \cite{yu2013harry} to set detections into non-overlapping groups to obey the constraint of mutual exclusion.

\subsubsection{Trajectory-level Exclusion Modeling}
Generally, trajectory-level exclusion is modeled by penalizing the case that two close detection hypotheses have different trajectory labels. This will suppress one trajectory label. For example, the penalty term in \cite{andriyenko2011multi} is inversely proportional to the distance between two detection responses with different trajectory labels. If two detection responses are too close, it will lead to a considerably large, or in the limit case, infinite, cost. \textcolor{blue}{A} similar idea is adopted in \cite{andriyenko2012discrete}. The penalty of trajectory-level exclusion in \cite{milan13detection} is proportional to the spatial-temporal overlap between two trajectories. The closer the two trajectories, the higher the penalty. There is also a special case \cite{Butt13multi}, in which the exclusion is modeled as an extra constraint to the so-called ``Conflict'' edges in a network flow based algorithm.


\subsection{Occlusion Handling}
\label{occlusion}
Occlusion is perhaps the most critical challenge in MOT. It is a primary cause for ID switches or fragmentation of trajectories. In order to handle occlusion, various kinds of strategies have been proposed.

\subsubsection{Part-to-whole}
This strategy is built on the assumption that a part of the object is still visible when an occlusion happens. This assumption holds in most cases. Based on this assumption, approaches adopting this strategy observe and utilize the visible part to infer the state of the whole object.

The popular way is dividing a holistic object (like a bounding box) into several parts and computing affinity based on individual parts. If an occlusion happens, affinities regarding occluded parts should be low. Tracker would be aware of this and adopt only the unoccluded parts for estimation. Specifically, parts are derived by dividing objects into grids uniformly \cite{hu2012single}, or fitting multiple parts into a specific kind of object like human, \eg 15 non-overlap parts as in \cite{yang2012onlinelearned}, and parts detected from the DPM detector \cite{felzenszwalb2010object} in \cite{izadinia2012,shu2012part}. 

Based on these individual parts, observations of the occluded parts are ignored. For instance, part-wise appearance model is constructed in \cite{hu2012single}. Reconstructed error is used to determine which part is occluded or not. The appearance model of the holistic object is selectively updated by only updating the unoccluded parts. This is the ``hard'' way of ignoring the occluded part, while there is a ``soft'' way in \cite{yang2012onlinelearned}. Specifically, the affinity concerning two tracklets $j$ and $k$ is computed as $\sum_i w_iF\left(\mathbf{f}_j^i,\mathbf{f}_k^i\right)$, where $\mathbf{f}$ is feature, $i$ is the index of parts. The weights are learned according to the occlusion relationship of parts. In \cite{izadinia2012}, human body part association is conducted to recover the part trajectory and further assists whole object trajectory recovery.

``Part-to-whole'' strategy is also applied in tracking based on feature point clustering, which assumes feature points with similar motion should belong to the same object. As long as some parts of an object are visible, the clustering of feature point trajectories will work \cite{sugimura2009using, brostow2006unsupervised, fragkiadaki2012two}.

\begin{figure}[t]
 \centering
 \includegraphics[width=0.99\linewidth]{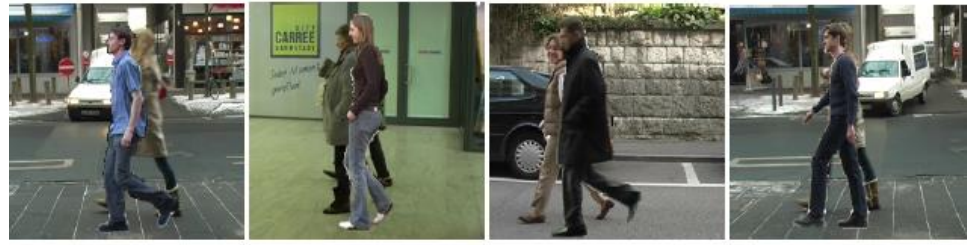}
 \vspace{-0.35cm}
 \caption{Training samples for the double-person detector \cite{tang2014detection}. From left to right, the level of occlusion increases}
 \label{fig:occlusion_ex4}
\end{figure}

\subsubsection{Hypothesize-and-test}

This strategy sidesteps challenges from occlusion by hypothesizing proposals and testing the proposals according to observations at hand. As the name indicates, this strategy is composed of two steps, \emph{hypothesize} and \emph{test}.

\textbf{Hypothesize.} Zhang \etal \cite{zhang2008global} generate occlusion hypotheses based on the occludable pair of observations, which are close and with similar scale. Assuming $\mathbf{o}_i$ is occluded by $\mathbf{o}_j$, a corresponding occlusion hypothesis is $\widetilde{\mathbf{o}}_i^j = \left(\mathbf{p}_j,s_i,\mathbf{f}_i,t_j\right)$, where $\mathbf{p}_j$ and $t_j$ are the position and time stamp of $\mathbf{o}_j$, and $s_i$ and $\mathbf{f}_i$ are the size and appearance feature of $\mathbf{o}_i$. This approach treats occlusion as distraction, while in other works \cite{tang2013learning, tang2014detection} occlusion patterns are employed to assist detection in case of occlusion. More specifically, different detection hypotheses are generated by synthetically combining two objects with different levels and patterns of occlusion (see Figure \ref{fig:occlusion_ex4}).

\textbf{Test.} The hypotheses would be employed for MOT when they are ready. Let us revisit the two approaches described above. In \cite{zhang2008global}, the hypothesized observations along with the original ones are given as input to the cost-flow framework and MAP is conducted to obtain the optimal solution. In \cite{tang2013learning} and \cite{tang2014detection}, a multi-person detector is trained based on the detection hypotheses. This detector greatly reduces the difficulty of detection in case of occlusion.

\subsubsection{Buffer-and-recover}
This strategy buffers observations when occlusion happens and remembers states of objects before occlusion. When occlusion ends, object states are recovered based on the buffered observations and the stored states before occlusion.

Mitzel \etal \cite{mitzel2010multi} keep a trajectory alive for up to 15 frames when occlusion happens, and extrapolates the position to grow the dormant trajectory through occlusion. In case the object reappears, the track is triggered again and the identity is maintained. This idea is followed in \cite{mitzel2011real}. Observation mode is activated when the tracking state becomes ambiguous due to occlusion \cite{ryoo2008observe}. As soon as enough observations are obtained, hypotheses are generated to explain the observations. This could also be treated as ``buffer-and-recover" strategy.

\subsubsection{Others}
The strategies described above may not cover all the tactics explored in the community. For example, Andriyenko \etal~\cite{andriyenko2011analytical} represent targets as Gaussian distributions in image space and explicitly model pair-wise occlusion ratios between all targets as part of a differentiable energy function. In general, it is non-trivial to distinctly separate or categorize various approaches to occlusion modeling, and in some cases, multiple strategies are used in combination.

\subsection{Inference}
\label{inference}

\subsubsection{Probabilistic Inference}
\label{prob_inference}
Approaches based on probabilistic inference typically represent states of objects as a distribution with uncertainty. The goal of a tracking algorithm is to estimate the probabilistic distribution of target state by a variety of probability reasoning methods based on existing observations. This kind of approaches typically requires only the existing, \ie past and present observations, thus they are especially appropriate for the task of online tracking. As only the existing observations are employed for estimation, it is natural to impose the assumption of Markov property in the objects state sequence. This assumption includes two aspects, recalling the formula in Section \ref{mot_formula}.

First, the current object state only depends on the previous states. Further, it only depends on the very last state if the first-order Markov property is imposed, which can be formalized as $P\left(\mathbf{S}_t|\mathbf{S}_{1:t-1}\right) = P\left(\mathbf{S}_t|\mathbf{S}_{t-1}\right)$.

Second, the observation of an object is only related to its state corresponding to this observation. In other words, the observations are conditionally independent: $P\left(\mathbf{O}_{1:t}|\mathbf{S}_{1:t}\right) = \prod_{i=1}^t P\left(\mathbf{O}_i|\mathbf{S}_i\right)$.

These two aspects are related to the \emph{Dynamic Model} and the \emph{Observation Model}, respectively. The dynamic model corresponds to the tracking strategy, while the observation model provides observation measurements concerning object states. The \emph{predict} step is to estimate the current state based on all the previous observations. More specifically, the posterior probability distribution of the current state is estimated by integrating in the space of the last object state via the dynamic model. The \emph{update} step is to update the posterior probability distribution of states based on the obtained measurements under the observation model.

According to the equations, states of objects can be estimated by iteratively conducting the prediction and updating steps. However, in practice, the object state distribution cannot be represented without simplifying assumptions, thus there is no analytical solution to computing the integral of the state distribution. Additionally, for multiple objects, the dimension of the sets of states is very large, which makes the integration even more difficult, requiring the derivation for approximate solutions.

Various kinds of probabilistic inference models haves been applied to multi-object tracking \cite{kratz2010tracking,fortmann1983sonar,giebel2004bayesian,ECCVW16MOTVBM}, such as Kalman filter \cite{rodriguez2011data,reid1979algorithm}, Extended Kalman filter \cite{mitzel2011real} and Particle filter \cite{jin2007variational,yang2005fast,hess2009discriminatively,han2007probabilistic,hu2012single,liu2012automatic,breitenstein2009robust,yang2009detection}.

\textbf{Kalman filter.} In the case of a linear system and Gaussian-distributed object states, the Kalman filter~\cite{reid1979algorithm} is proven to be the optimal estimator. It has been applied in \cite{rodriguez2011data}.

\textbf{Extended Kalman filter.} To include the non-linear case, the extended Kalman filter is one possible solution. It approximates the non-linear system by a Taylor expansion \cite{mitzel2011real}.

\textbf{Particle filter.} Monte Carlo sampling based models have also become popular in tracking, especially after the introduction of the particle filter \cite{jin2007variational, yang2005fast, hess2009discriminatively, hu2012single, liu2012automatic, breitenstein2009robust, yang2009detection, khan2004mcmc}. This strategy models the underlying distribution by a set of weighted particles, thereby allowing to drop any assumptions about the distribution itself \cite{liu2012automatic, breitenstein2009robust, yang2009detection, kratz2010tracking}.


\subsubsection{Deterministic Optimization}
\label{deterministic}
As opposed to the probabilistic inference methods, approaches based on deterministic optimization aim to find the maximum a posteriori (MAP) solution to MOT. To that end, the task of inferring data association, the target states or both, is typically cast as an optimization problem. Approaches within this framework are more suitable for the task of offline tracking because observations from all the frames or at least a time window are required to be available in advance. Given observations (usually detection hypotheses) from all the frames, these types of methods endeavor to globally associate observations belonging to an identical object into a trajectory. The key issue is how to find the optimal association. Some popular and well-studied approaches are detailed in the following.

\textbf{Bipartite graph matching.} By modeling the MOT problem as bipartite graph matching, two disjoint sets of graph nodes could be existing trajectories and new detections in online tracking or two sets of tracklets in offline tracking. Weights among nodes are modeled as affinities between trajectories and detections. Then either a greedy bipartite assignment algorithm \cite{shu2012part, breitenstein2009robust, wu2007detection} or the optimal Hungarian algorithm \cite{qin2012improving, reilly2010detection, perera2006multi, xing2009multi, huang2008robust} are employed to determine the matching between nodes in the two sets.

\textbf{Dynamic Programming.} Extend dynamic programming \cite{wolf1989finding}, linear programming \cite{jiang2007linear, berclaz2009multiple, andriyenko2010globally}, quadratic boolean programming \cite{leibe2007coupled}, K-shortest paths \cite{berclaz2011multiple,choi2012unified}, set cover \cite{wu2011efficient} and subgraph multicut \cite{tang2015subgraph,ECCVW16MultiCut,Tang_2017_CVPR}, maximum multi-clique \cite{Dehghan_2015_CVPR2} are adopted to solve the association problem among detections or tracklets.


\textbf{Min-cost max-flow network flow.} Network flow is a directed graph where each edge has a certain capacity. For MOT, nodes in the graph are detection responses or tracklets. Flow is modeled as an indicator to link two nodes or not. To meet the flow balance requirement, a source node and a sink node corresponding to the start and the end of a trajectory are added to the graph. One trajectory corresponds to one flow path in the graph. The total flow transited from the source node to the sink node equals to the number of trajectories, and the cost of transition is the negative log-likelihood of all the association hypotheses. Note that the globally optimal solution can be obtained in polynomial time, \eg using the push-relabel algorithm. This model is exceptionally popular and has been widely adopted~\cite{zhang2008global,choi2012unified,wu2012coupling,Butt13multi,pirsiavash2011globally,Lenz:2015:ICCV,Dehghan_2015_CVPR,Chari_2015_CVPR,Gaidon_2016_CVPR}.
	
\textbf{Conditional random field.} The conditional random field model is adopted to handle the MOT problem in \cite{yang2012online,yang2011learning,milan13detection,ECCVW16LTTS}. Defining a graph $G = \left(V,E\right)$ where $V$ is the set of nodes and $E$ is the set of edges, low-level tracklets are given as input to the graph. Each node in the graph represents observations~\cite{milan13detection} or pairs of tracklets~\cite{yang2012online}, and a label is predicted to indicate which track the observations belongs to or whether to link the tracklets. 
	
\textbf{MWIS.} The maximum-weight independent set (MWIS) is the heaviest subset of non-adjacent nodes of an attributed graph. As in the CRF model described above, nodes in the attribute graph represent pairs of tracklets in successive frames, weights of nodes represent the affinity of the tracklet pair, and the edge is connected if two tracklets share the same detection. Given this graph, the data association is modeled as the MWIS problem \cite{shafique2008rank,brendel2011multiobject}.

\subsubsection{Discussion}
\label{inference_discuss}
In practice, deterministic optimization or energy minimization is employed more popularly compared with probabilistic approaches. Although the probabilistic approaches provide a more intuitive and complete solution to the problem, they are usually difficult to infer. On the contrary, energy minimization could obtain a ``good enough'' solution in a reasonable time.

\subsection{Summary}
As described above, we have introduced and reviewed different components of an MOT system. It is important to note that not all existing MOT methods have all the components. For example, interaction is not modeled in some studies. Some models are only necessary in specific cases, such as the crowd motion pattern in the case of extremely crowded scenarios. Occlusion is not specifically handled in some of the existing works. In general, appearance, motion and inference are mandatory in most methods. Let us take the simplest case as an example, \emph{i.e.}, using a single tracker to track each object individually. In this example, interaction, exclusion and occlusion are not addressed. But appearance and motion models are still necessary with an inference model.

It is also notable that, these components are not orthogonal to each other. They can usually be combined and integrated for satisfactory performance. For instance, the interaction is modeled as several terms along with terms regarding appearance, motion and exclusion modeling, and the resulted objective is optimized with deterministic techniques \cite{yamaguchi2011you}. Four kinds of features like appearance, motion and location features are concatenated for computing tracklet-object similarity with a two-layer network in \cite{Xu_2019_ICCV}. Appearance and motion features are fused by affinity sub-net for more powerful discrimination in \cite{Chu_2019_ICCV}.

\section{MOT Evaluation}
\label{mot_evaluation}
For a given MOT approach, metrics and datasets are required to evaluate its performance quantitatively. This is important for two reasons. On the one hand, it is essential to measure the influence of different components and parameters on the overall performance to design the best system. On the other hand, it is desirable to have a direct comparison to other methods. Performance evaluation for MOT is not straightforward, as we will see in this section.

Evaluation metrics \cite{leichter2013monotonicity} of MOT approaches are crucial as they provide a means for fair quantitative comparison. A brief review on different MOT evaluation metrics is presented in this section. As many approaches to MOT employ the tracking-by-detection strategy, they often measure detection performance as well as tracking performance. Metrics for object detection are therefore adopted in MOT approaches. Based on this, MOT metrics can be roughly categorized into two sets evaluating detection and tracking respectively, as listed in Table~\ref{table:metrics}.

\subsection{Metrics}
\label{mot_metrics}

\begin{table}[h]
\scriptsize
\caption{An overview of evaluation metrics for MOT. The up arrow (\textit{resp}. down arrow) indicates that the performance is better if the quantity is greater (\textit{resp}. smaller).}
\vspace{-0.35cm}
\tabcolsep=2pt
\label{table:metrics}
\centering
{\begin{tabular}{ccc|ccc}
\hline
\textbf{Metric} & \textbf{Description} & \textbf{Note} & \textbf{Metric} & \textbf{Description} & \textbf{Note} \\
\hline
Recall& \tabincell{l}{Ratio of correctly matched detections \\to ground-truth detections} & $\uparrow$  & TDE& \tabincell{l}{Distance between the ground-truth\\ annotation and the tracking result} & $\downarrow$\\
\hline
Precision& \tabincell{l}{Ratio of correctly matched \\detections to total\\ result detections} & $\uparrow$  & OSPA& \tabincell{l}{Cardinality error, label error\\ and spatial distance between \\ground truth and the tracking results} & $\downarrow$ \\
\hline
FAF/FPPI& \tabincell{l}{Number of false alarms per frame\\averaged over a sequence} & $\downarrow$  &  MT & \tabincell{l}{Percentage of ground-truth trajectories\\ which are covered by the tracker \\output for more than 80\% of their length} & $\uparrow$\\
\hline

MODA& \tabincell{l}{Combines missed detections \\and FAF} & $\uparrow$  & ML&\tabincell{l}{Percentage of ground-truth trajectories \\which are covered by the tracker \\output for less than 20\% of their length} & $\downarrow$\\
\hline

MODP& \tabincell{l}{Average overlap between true \\positives and ground truth} & $\uparrow$  & PT & \tabincell{l}{1.0 - MT - ML} & -\\
\hline

MOTA & \tabincell{l}{Combines false negatives, false \\positives and mismatch rate} & $\uparrow$   & FM& \tabincell{l}{Number of times that a ground-truth\\ trajectory is interrupted in the\\ tracking result} & $\downarrow$\\
\hline
IDS& \tabincell{l}{Number of times that a tracked\\ trajectory changes its matched\\ ground-truth identity\\ (or vice versa)} & $\downarrow$  & RS& \tabincell{l}{Ratio of tracks which are \\correctly recovered from short occlusion} & $\uparrow$ \\
\hline

MOTP& \tabincell{l}{Overlap between the estimated \\positions and the ground truth \\averaged over the matches} & $\uparrow$  & RL& \tabincell{l}{Ratio of tracks which are \\correctly recovered from long occlusion} & $\uparrow$\\
\hline

\end{tabular}}
\end{table}

\subsubsection{Metrics for Detection}
We further group metrics for detection into two subsets. One set measures accuracy, and the other one measures precision.

\textbf{\textit{Accuracy.}} The commonly used Recall and Precision metrics as well as the average False Alarms per Frame (FAF) rate are employed as MOT metrics~\cite{yang2011learning}. Choi \etal \cite{choi2010multiple} use the False Positive Per Image (FPPI) to evaluate detection performance in MOT. A comprehensive metric called Multiple Object Detection Accuracy (MODA), which considers the relative number of false positives and miss detections is proposed in \cite{kasturi2009framework}.
	
\textbf{\textit{Precision.}} The Multiple Object Detection Precision (MODP) metric measures the quality of alignment between predicted detections and the ground truths \cite{kasturi2009framework}.

\subsubsection{Metrics for Tracking} Metrics for tracking are classified into four subsets by different attributes as follows.

\textbf{\textit{Accuracy.}} This kind of metrics measures how accurately an algorithm can track targets. The metric of ID switches (IDs)~\cite{yamaguchi2011you} counts how many times an MOT algorithm switches between objects. The Multiple Object Tracking Accuracy (MOTA) metric \cite{keni2008evaluating} combines the false positive rate, false negative rate and mismatch rate into a single number, giving a fairly reasonable quantity for the overall tracking performance. Despite some drawbacks and criticisms, this is by far the most widely accepted evaluation measure for MOT.

\textbf{\textit{Precision.}} Three metrics, Multiple Object Tracking Precision (MOTP)~\cite{keni2008evaluating}, Tracking Distance Error (TDE)~\cite{kratz2010tracking} and OSPA \cite{ristic2011metric} belong to this subset. They describe how precisely the objects are tracked measured by bounding box overlap and/or distance. Specifically, cardinality error and label error are additionally considered in \cite{ristic2011metric}.

\textbf{\textit{Completeness.}} Metrics for completeness indicate how completely the ground truth trajectories are tracked. The numbers of Mostly Tracked (MT), Partly Tracked (PT), Mostly Lost (ML) and Fragmentation (FM)~\cite{li2009learning} belong to this set.

\textbf{\textit{Robustness.}} To assess the ability of an MOT algorithm to recover from occlusion, metrics called Recover from Short-term occlusion (RS) and Recover from Long-term occlusion (RL) are introduced in~\cite{song2010stochastic}.

\subsection{Datasets}
\label{datasets}

\begin{table}[t]
\scriptsize
\caption{Statistics of publicly available datasets. \# V and \# F mean how many videos and frames are included in the dataset. ``Multi-view'' and ``GT'' indicate whether multi-view data and ground truth are provided. ``Indoor'' and ``Outdoor'' denote the types of scenarios in the dataset. For the last four items, the check and cross marks mean YES and NO, respectively. The page of a dataset can be accessed by clicking the dataset name.}
 \vspace{-0.35cm}
\label{table:datasets}
\centering
{\begin{tabular}{lclclclclclclc}
\hline
\textbf{Dataset} & \textbf{\# V} & \textbf{\# F} & \tabincell{l}{\textbf{Multi-} \\ \textbf{view}}  & \textbf{GT}& \textbf{Indoor} & \textbf{Outdoor}\\
\hline
\href{https://motchallenge.net/data/MOT16}{MOT16}  &  14 & 11K & \xmark & \checkmark& \checkmark & \checkmark \\
\hline
\href{www.cvlibs.net/datasets/kitti/eval_tracking.php}{KITTI} & 50 & -- & \checkmark &  \checkmark & \xmark &  \checkmark\\
\hline
\href{http://www.cvg.reading.ac.uk/PETS2016/}{PETS 2016}  & 13 & -- & \checkmark & \checkmark & \xmark & \checkmark\\
\hline
\href{http://www.cvg.reading.ac.uk/PETS2009/a.html}{PETS 2009}  & 3 & -- & \checkmark & \checkmark & \xmark &  \checkmark \\
\hline
\href{http://groups.inf.ed.ac.uk/vision/CAVIAR/CAVIARDATA1/}{CAVIAR}  & 54 & -- &\checkmark & \checkmark   & \checkmark & \xmark\\
\hline
\href{https://www.mpi-inf.mpg.de/departments/computer-vision-and-multimodal-computing/software-and-datasets/}{TUD Stadtmitte}  & 1 & 179 & \xmark  & \checkmark & \xmark &  \checkmark\\
\hline
\href{https://www.mpi-inf.mpg.de/departments/computer-vision-and-multimodal-computing/software-and-datasets/}{TUD Campus}  & 1 & 71 & \xmark  & \checkmark & \xmark &  \checkmark\\
\hline
\href{https://www.mpi-inf.mpg.de/departments/computer-vision-and-multimodal-computing/software-and-datasets/}{TUD Crossing}  & 1 & 201 & \xmark  & \checkmark & \xmark &  \checkmark\\

\hline
\href{https://www.vision.caltech.edu/Image_Datasets/CaltechPedestrians/}{Caltech Pedestrian} &137 & 250K& \xmark  & \checkmark &  \xmark &  \checkmark\\
\hline
\href{www.cs.ubc.ca/~okumak/research.html}{UBC Hockey}  & 1 & $\approx$100& \xmark  & \xmark & \xmark &  \checkmark\\
\hline
\href{https://data.vision.ee.ethz.ch/cvl/aess/dataset/}{ETH Pedestrian}  & 8 & 4K & \checkmark & \checkmark & \xmark & \checkmark\\
\hline
\href{http://www.vision.ee.ethz.ch/datasets/}{ETHZ Central}  & 3 & 13K & \xmark & \checkmark & \xmark & \checkmark \\
\hline
\href{www.robots.ox.ac.uk/ActiveVision/Research/Projects/2009bbenfold_headpose/project.html\#datasets}{Town Centre}  &1 & 4.5K&\xmark & \checkmark &  \xmark &  \checkmark\\ 
\hline
\href{https://graphics.cs.ucy.ac.cy/research/downloads/crowd-data}{Zara}  & 4 & -- & \xmark & \xmark &  \xmark &  \checkmark\\
\hline
\href{http://www.svcl.ucsd.edu/projects/anomaly/dataset.htm}{UCSD}  & 98 & -- & \xmark & \xmark &  \xmark & \checkmark\\
\hline
\href{http://www.crcv.ucf.edu/data/tracking.php}{UCF Marathon}  & 3 & 1.3K & \xmark & \checkmark &  \xmark & \checkmark\\
\hline
\href{http://www.crcv.ucf.edu/data/ParkingLOT/}{ParkingLOT}  & 3 & 2.7K & \xmark & \checkmark  & \xmark & \checkmark\\

\hline
\end{tabular}}
\end{table}

To compare with various existing methods and determine the state of the arts in MOT, publicly available datasets are employed to evaluate the proposed methods in individual publications. Table~\ref{table:datasets} gives the most popular datasets used in the literature and provides detailed statistics of these datasets.

These datasets have played an important role in the progress of MOT. However, there are some issues with them. First, the scale of datasets for MOT is relatively smaller than that of SOT, \eg, the sequences used in the online object tracking benchmark \cite{wu2013online} and the VOT challenge \cite{VOT_TPAMI}, which drive the fast development and standardized evaluation of SOT. Second, current datasets focus on pedestrians. This can be attributed to the fact that pedestrian detection has achieved great success in recent years. However, exciting progress of multi-class detection has been achieved in more recent years. We believe multi-class-multi-object tracking is feasible building upon the detection module of multi-class objects. Thus, it is time to move towards datasets of multi-class objects for MOT. 

\subsection{Public Algorithms}
\label{pub_algorithms}

\begin{table}[t]
\scriptsize
\caption{A list of publicly available program codes. Please click the reference to access the corresponding code.}
 \vspace{-0.35cm}
\label{table:algorithms}
\tabcolsep=12pt
\centering
\scriptsize
{\begin{tabular}{ l l |l l }
\hline
 \textbf{Ref.} & \textbf{Note} & \textbf{Ref.} & \textbf{Note}\\
\hline
\href{https://www.eecs.umich.edu/vision/mttproject.html}{Choi \& Savarese~\cite{choi2010multiple}} & C++ &  \href{http://www.cs.cmu.edu/~deva/pubs.html/}{Pirsiavash \etal~\cite{pirsiavash2011globally}} & MATLAB\\

\href{http://www.cs.bc.edu/~hjiang/details/tracking/}{Jiang \etal~\cite{jiang2007linear}} &  C  &  \href{https://lrs.icg.tugraz.at/download.php#motog}{Possegger \etal~\cite{possegger2014occlusion}} & MATLAB \\
\href{http://research.milanton.de/contracking/}{Milan \etal~\cite{milan2014continuous}} &  MATLAB &  \href{http://www.mikelrodriguez.com/crowd-tracking-matlab-application}{Rodriguez \etal~\cite{rodriguez2009tracking}} & MATLAB  \\
\href{http://research.milanton.de/dctracking/}{Andriyenko \etal~\cite{andriyenko2012discrete}} & MATLAB  & \href{https://github.com/abewley/sort}{Bewley \etal~\cite{Bewley2016_sort}} & Python \\
\href{https://bitbucket.org/amilan/dctracking/}{Milan \etal~\cite{milan13detection}} &  MATLAB &  \href{http://rehg.org/mht/}{Kim \etal~\cite{kim_ICCV2015_MHTR}} & MATLAB \\
\href{http://crcv.ucf.edu/projects/GMCP-Tracker/}{Zamir \etal~\cite{zamir2012gmcp}} & MATLAB &  \href{https://bitbucket.org/cdicle/smot}{Dicle \etal~\cite{dicle2013way}} & MATLAB  \\
\href{http://cvlab.epfl.ch/software/ksp/}{Berclaz \etal~\cite{berclaz2011multiple}} &  C &  \href{https://cvl.gist.ac.kr/cmot.html}{Bae \& Yoon~\cite{bae2014robust}} & MATLAB   \\
\href{http://www.milanton.de/m-best}{Rezatofighi \etal~\cite{Rezatofighi:2015:ICCV}} & MATLAB &  \href{https://github.com/yuxng/MDP_Tracking}{Xiang \etal~\cite{xiang2015learning}} & MATLAB  \\

\href{http://visionlab.tudelft.nl/spot/}{Zhang \& Laurens~\cite{zhang2013structure,zhang2013preserving}} & MATLAB/C &  \href{https://github.com/francescosolera/LDCT}{Solera \etal~\cite{solera2015learning}} & MATLAB \\

\hline
\end{tabular}}
\end{table}

We examine the literature and list algorithms for which the associated source codes are publicly available to make further comparisons convenient in Table~\ref{table:algorithms}.

Compared with SOT, there seem to be not many public programs. Admittedly, progress in SOT is larger than that of MOT recently. One reason can be that, many researchers have made their codes publicly available. We here encourage researchers to publish code for research convenience of others in the future.

\begin{table*}[t]
\centering
\caption{Quantitative results on the PETS2009-S2L1 dataset, extracted from the respective publications. Methods, ordered by year of publication, are labeled with Online and Offline tags. Concerning each metric, the best results are shown in bold. The table shows that progress has been achieved while there seems not much research attention in this topic recently.}
 \vspace{-0.35cm}
\label{table:benchmark_pets09S2L1}
\tabcolsep=1.8pt
\centering
\tiny
{\begin{tabular}{l|c|c|c|c|c|c|c|c|c|c|c|c|r}
\hline
\textbf{Ref.} & \textbf{MOTA} $\uparrow$ & \textbf{MOTP} $\uparrow$ & \textbf{IDS} $\downarrow$ & \textbf{Pre.} $\uparrow$ & \textbf{Rec.} $\uparrow$ & \textbf{MT} $\uparrow$ & \textbf{PT} & \textbf{ML} $\downarrow$ & \textbf{FM} $\downarrow$ & \textbf{F1} $\uparrow$ & \textbf{Year} & \textbf{Note} &\textbf{Source}\\
\hline

Berclaz \etal~\cite{berclaz2009multiple} & 0.830 & 0.520 & - & 0.820 & 0.530 & - & - & - & - & 0.644 & 2009 & Off & \cite{izadinia2012}\\

Yang \etal~\cite{yang2009probabilistic} & 0.759 & 0.538 & - & - & - & - & - & - & - & - & 2009 & On & \cite{possegger2014occlusion}\\

Alahi \etal~\cite{alahi2009sparsity} & 0.830 & 0.520 & - & 0.690 & 0.530 & - & - & - & - & 0.600 & 2009 & On & \cite{izadinia2012} \\

\hline
Conte \etal~\cite{conte2010performance} & 0.810 & 0.570 & - & 0.850& 0.580 & - & - & - & - & 0.690 & 2010 & On & \cite{izadinia2012} \\

\hline

Berclaz \etal~\cite{berclaz2011multiple} & 0.803 & 0.720 & 13 & 0.963 & 0.838 & 0.739 & 0.174 & 0.087 & 22 & 0.896 &2011 & Off &\cite{wen2014multiple}\\

Shitrit \etal~\cite{ben2011tracking} & 0.815 & 0.584 & 19 & 0.907 & 0.908 & - & - & - & - & 0.907 & 2011 & Off & \cite{zamir2012gmcp}\\

Andriyenko \& Schindler~\cite{andriyenko2011multi} & 0.863 & 0.787 & 38 & 0.976 & 0.895 & 0.783 & 0.174 & 0.043 & 21 & 0.934 & 2011 & Off & \cite{wen2014multiple}\\

Henriques \etal~\cite{henriques2011globally} & 0.848 & 0.687 & 10 & 0.924 & 0.940 & - & - & - & - & 0.932 & 2011 & Off & \cite{zamir2012gmcp}\\

Pirsiavash \etal~\cite{pirsiavash2011globally} & 0.774 & 0.743 & 57 & 0.972 & 0.812 & 0.609 & 0.347 & 0.043 & 62 & 0.885 & 2011 & Off & \cite{wen2014multiple}\\

Kuo \etal~\cite{kuo2011does} & - & - & 1 & \textbf{0.996} & 0.895 & 0.789 & 0.211 & \textbf{0.000} & 23 & 0.943 & 2011 & Off & \cite{yang2012multi}\\

Andriyenko \etal~\cite{andriyenko2011analytical} & 0.917 & 0.745 & 11 & - & - & - & - & - & - & - & 2011 & Off & \cite{zhang2012online}\\

Leal \etal~\cite{leal2011everybody} & 0.670 & - & - & - & - & - & - & - & - & - &2011 & Off & \cite{zhang2012online}\\

Breitenstein \etal~\cite{breitenstein2011online} & 0.797 & 0.563 & - & - & - & - & - & - & - & - & 2011 & On & \cite{possegger2014occlusion}\\

\hline

Izadinia \etal~\cite{izadinia2012} & 0.907 & 0.760 & - & 0.968 & 0.952 & - & - & - & - & 0.960 & 2012 & Off & \cite{izadinia2012}\\

Zamir \etal~\cite{zamir2012gmcp} & 0.903 & 0.690 & 8 & 0.936 & 0.965 & - & - & - & - & 0.950 & 2012 & Off & \cite{zamir2012gmcp}\\

Andriyenko \etal~\cite{andriyenko2012discrete} & 0.883 & 0.796 & 18 & 0.987 & 0.900 & 0.826 & 0.174 & \textbf{0.000} & 14 & 0.941 & 2012 & Off & \cite{wen2014multiple}\\

Yang \& Nevatia~\cite{yang2012multi} & - & - & \textbf{0} & 0.990 & 0.918 & 0.895 & 0.105 & \textbf{0.000} & 9 & 0.953 & 2012 & Off & \cite{yang2012multi}\\

Yang \& Nevatia~\cite{yang2012onlinelearned} & - & - & \textbf{0} & 0.948 & 0.978 & 0.950 & 0.050 & \textbf{0.000} & \textbf{2} & 0.963 & 2012 & Off & \cite{zhang2015multi}\\

Leal \etal~\cite{leal2012branch} & 0.760 & 0.600 & - & - & - & - & - & - & - & - &2012 & Off & \cite{possegger2014occlusion}\\

Zhang \etal~\cite{zhang2012online} & 0.933 & 0.682 & 19 & - & - & - & - & - & - & - & 2012 & On & \cite{zhang2012online}\\

\hline

Segal \& Reid~\cite{segal2013latent} & 0.900 & 0.750 & 6 & - & - & 0.890 & - & - & - & - & 2013 & Off & \cite{dehghan2015target}\\

Kumar \& Vleeschouwer~\cite{kc2013discriminative} & 0.910 & 0.700 & 5 & - & - & - & - & - & - & - & 2013 & Off & \cite{dehghan2015target}\\

Hofman \etal~\cite{hofmannhypergraphs} & 0.980 & 0.828 & 10 & - & - & \textbf{1.000} & 0.000 & \textbf{0.000} & 11 & - & 2013 & Off & \cite{possegger2014occlusion}\\

Milan \etal~\cite{milan13detection} & 0.903 & 0.743 & 22 & - & - & 0.783 & 0.217 & \textbf{0.000} & 15 & - & 2013 & Off & \cite{milan13detection}\\

Hofmann \etal~\cite{hofmann2013unified} & 0.978 & 0.753 & 8 & 0.991 & \textbf{0.990} & \textbf{1.000} & 0.000 & \textbf{0.000} & 8 & \textbf{0.990} & 2013 & Off & \cite{hofmann2013unified}\\

Shi \etal~\cite{shi2013multi} & 0.927 & 0.818 & 7 & 0.982 & 0.960 & 0.947 & 0.053 & \textbf{0.000} & 11 & 0.971 & 2013 & Off & \cite{shi2014multi}\\

Wu \etal~\cite{wu2013onlinemotion} & 0.928 & 0.743 & 8 & - & - & \textbf{1.000} & 0.000 & \textbf{0.000} & 11 & - & 2013 & On & \cite{wu2013onlinemotion} \\

\hline

Milan \etal~\cite{milan2014continuous} & 0.906 & 0.802 & 11 & 0.984 & 0.924 & 0.913 & 0.043 & 0.043 & - & 0.953 & 2014 & Off & \cite{wen2014multiple}\\

Wen \etal~\cite{wen2014multiple} & 0.927 & 0.729 & 5 & 0.984 & 0.944 & 0.957 & 0.043 & \textbf{0.000} & 10 & 0.964 & 2014 & Off & \cite{wen2014multiple}\\

Shi \etal~\cite{shi2014multi} & 0.961 & 0.818 & 4 & 0.989 & 0.977 & 0.947 & 0.053 & \textbf{0.000} & 6 & 0.983 & 2014 & Off & \cite{shi2014multi}\\

Bae \& Yoon~\cite{bae2014robust} & 0.830 & 0.696 & 4 & - & - & \textbf{1.000} & 0.000 & \textbf{0.000} & 4 & - & 2014 & On & \cite{bae2014robust}\\

Possegger \etal~\cite{possegger2014occlusion} & \textbf{0.981} & 0.805 & 9 & - & - & \textbf{1.000} & 0.000 & \textbf{0.000} & 16 & - & 2014 & On & \cite{possegger2014occlusion}\\

\hline

Zhang \etal~\cite{zhang2015multi} & 0.956 & \textbf{0.916} & \textbf{0} & 0.986 & 0.970 & 0.950 & 0.050 & \textbf{0.000} & 4 & 0.978 & 2015 & Off & \cite{zhang2015multi}\\

Dehghan \etal~\cite{dehghan2015target} & 0.904 & 0.631 & 3 & - & - & 0.950 & 0.050 & \textbf{0.000} & - & - & 2015 & Off & \cite{dehghan2015target}\\

Lenz \etal \cite{Lenz:2015:ICCV} & 0.890 & 0.870 & 7 & - & - & 0.890 & 0.110 & 0 & 100 & - & 2015 & On &  \cite{Lenz:2015:ICCV}\\

\hline
\end{tabular}}
\end{table*}

\subsection{Benchmark Results}
\label{benchmark_results}
We list public results on the datasets mentioned above to get a direct comparison among different approaches and provide convenience for future comparison. Due to space limitation, we only present results on the most commonly employed \emph{PETS2009-S2L1} sequence in Table \ref{table:benchmark_pets09S2L1}. Results of other datasets are present in the supplementary material. Please note that this kind of direct comparison on the same dataset may not be fair due to the following points:
\vspace{-0.3cm}
\begin{itemize}\setlength{\itemsep}{-0.2cm}
	\item Different methodologies. For example, some publications belong to offline methods while others belong to online ones. Due to the difference described in Section \ref{online_vs_offline}, it is unfair to directly compare them because the former have access to much more information.
	\item Different detection hypotheses. Different approaches adopt various detectors to obtain detection hypotheses. One approach based on different detection hypotheses would output different results, let alone different approaches.
	\item Some approaches aggregate observations from multiple views while others utilize information from a single view. This makes a direct comparison between them difficult.
	\item Prior information, such as scene structure and the number of pedestrians, are exploited by some approaches, making a direct quantitative comparison to other methods which do not use that information questionable.
\end{itemize}
\vspace{-0.3cm}

\begin{table}[t]
\scriptsize
\caption{Benchmark result comparison between offline and online methods on the PETS2009-S2L1 dataset.}
 \vspace{-0.35cm}
\label{table:benchmark_comparison_pets2009}
\tabcolsep=8pt
\centering
{\begin{tabular}{lrrrrrr}
\toprule
\textbf{Proc.} &
\textbf{MOTA}$\uparrow$ & \textbf{MOTP}$\uparrow$ & \textbf{MT}$\uparrow$ & \textbf{ML}$\downarrow$ & \textbf{IDS}$\downarrow$ & \textbf{FM}$\downarrow$\\
\midrule
offline &
0.935 & 0.775 & 0.95 & 0 & 5.9 & 8.3 \\
online &
 0.913 & 0.748 & 1.00 & 0 & 7.0 & 10.3 \\
\bottomrule
\end{tabular}}
\end{table}

Strictly speaking, in order to make a direct and fair comparison, one needs to fix all the other components while varying the one under consideration. For instance, adopting different data association models while keeping all other parts the same could directly compare performance of different data association methods. This is the main goal of recent MOT benchmarks like KITTI~\cite{Geiger2012CVPR} and MOTChallenge~\cite{leal2015motchallenge, MOT16}, which specifically focus on a centralized evaluation of multiple object tracking. For intensive experimental comparison among different MOT solutions, please refer the respective benchmarks. In spite of the issues mentioned above, it is still worthy to list all the public results on the same dataset or sequence due to the following reasons.
\vspace{-0.3cm}
\begin{itemize}\setlength{\itemsep}{-0.2cm}
	\item By compiling all published results into a single table, it at least provides an intuitive comparison among different methods on the same dataset and a convenience for future works.
	\item Although this particular comparison among individual methods may not be fair, an approximate comparison between different types of methods such as that between offline and online methods could tell us how these types of methods perform on public datasets.
	\item Additionally, we could observe how the research of MOT progresses over time by comparing performance of methods across years.
\end{itemize}
\vspace{-0.3cm}
We report the results in terms of the MOTA, MOTP, IDS, Precision, Recall, MT, PT, ML, FM, and F1 metrics. At the same time, we tag the results with online and offline labels. Please note that: 1) there are missing entries, because we did not find the corresponding value neither from the original publication nor from other publications which cite it, and 2) in some cases, there could be different results for a unique publication (for example, results from the original publication versus results from another publication which compares with it). This discrepancy may arise because different configurations are adopted (\eg different detection hypotheses). In this case, we quote the most popularly cited one.



We conduct an analysis of benchmark results on the PETS2009-S2L1 dataset to investigate the comparison between offline methods and online methods. Choosing results from publications that appeared in 2012 and later, we average values of each metric across each type of methods, and report the mean value in Table \ref{table:benchmark_comparison_pets2009}. As expected, offline methods generally outperform online ones \emph{w.r.t.} most metrics. This is due to the fact that offline methods employ global temporal information for the estimation.

\begin{figure*}[t!]
\centering
  \begin{subfigure}[t]{0.48\textwidth}
	\centering
  \includegraphics[height=1.5in]{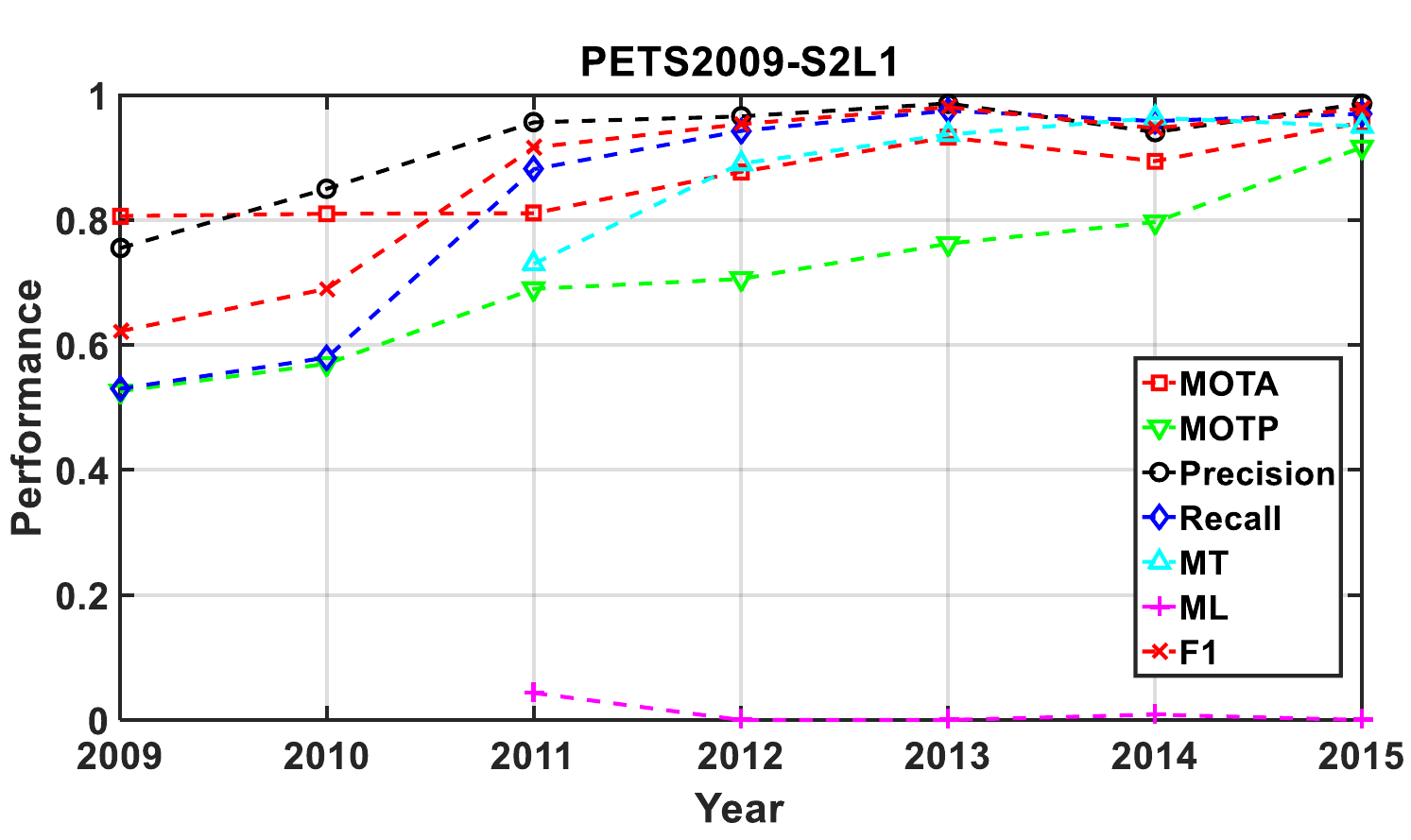}
  \end{subfigure}
	~
  \begin{subfigure}[t]{0.48\textwidth}
	\centering
   \includegraphics[height=1.5in]{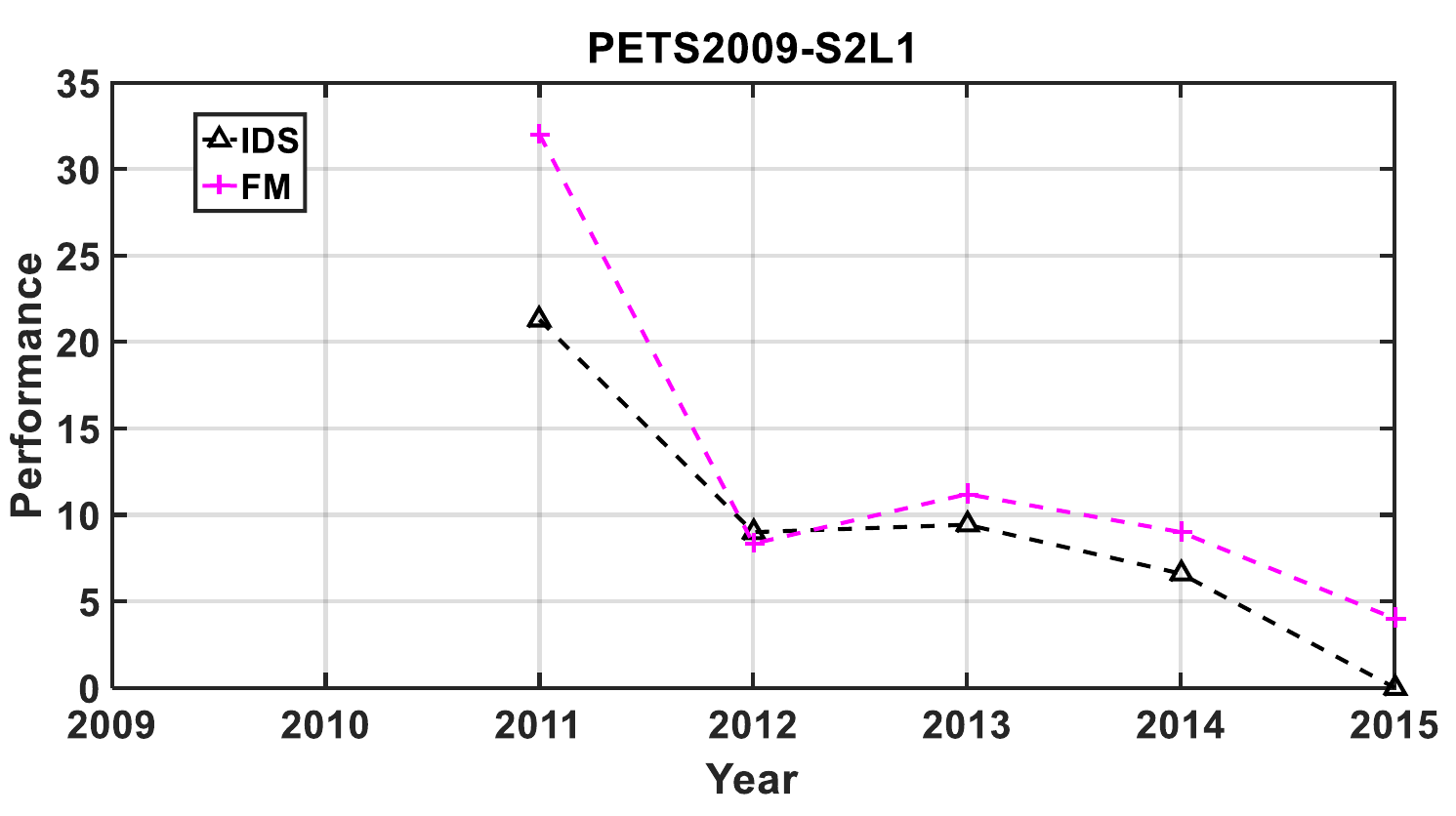}
  \end{subfigure}
	\scriptsize
	 \vspace{-0.35cm}
\caption{Statistics of results in different years on the PETS2009-S2L1 dataset in terms of MOTA, MOTP, Precision, Recall, MT, ML, F1 metrics (left), IDS and FM metrics (right).}
\label{fig:year_analysis}
\end{figure*}

Additionally, we analyze the evaluation results on the PETS2009-S2L1 dataset over time. Specifically, we plot the metric values of methods in each year ranging from 2009 to 2015 in Figure \ref{fig:year_analysis}. It is no surprise that the performance improves over the years. We suspect that factors such as better models and progress in object detection \cite{CVPR14RCNN, CVPR15FastRCNN, NIPS15FasterRCNN, CVPR16YOLO,ECCV16SSD,Hu_2018_CVPR,dai2016r,Dai_2017_ICCV} all contribute to the achieved progress. It should also be noted that a research community focuses on a specific dataset over time and certain methods may be a result of ``over-fitting'' to that dataset as opposed to general progress towards solving the problem.

\section{Summary}
\label{sec_summary}
This paper has described methods and problems related to the task of Multiple Object Tracking (MOT) in videos. As the first comprehensive literature review in the past decade, it has presented a unified problem formulation and several ways of categorization of existing methods, described the key components within state-of-the-art MOT approaches, and discussed the evaluations of MOT algorithms including evaluation metrics, public datasets, open source implementations, and benchmark results. Although great progress in MOT has been made in the past decades, there still remain several issues in the current MOT research and many open problems to be studied.

\subsection{Existing Issues}
\label{ssec_issues}
We have discussed the existing issues of datasets (Section \ref{datasets}) and public algorithms (Section \ref{pub_algorithms}). Except these issues, there are still some remarkable ones as follows:

One major issue in the MOT research is that, performance of an MOT method depends heavily on the object detectors. For example, the widely used tracking-by-detection paradigm is built upon an object detector, which provides detection hypotheses to drive the tracking procedure. Given different sets of detection hypotheses while fixing other components, an identical approach would produce tracking results with significant performance differences. In the community, sometimes no description about the detection module is given in the approach. This makes the comparisons with other approaches infeasible. Established benchmarks like KITTI and MOTChallenge attempt to alleviate this problem and are also moving towards a more principled and unified evaluation of detection and tracking (cf. MOT17).

Another nuisance is that, when developing an MOT solution, there are many parameters if this algorithm is too complicated. This leads to a difficulty of tuning the method. Meanwhile, it is also difficult for others to implement the approach and reproduce the result.

Some approaches perform well in specific video sequences. While applied to other cases, however, they may not produce satisfying results. The reasons are multi-fold. Differences of camera view, or the status of camera (moving versus static) can lead to this issue. It might also be caused by the fact that object detectors utilized by MOT approaches are trained in specific videos and do not generalize well in other video sequences.

All these issues restrict further development of the MOT research and its applications in practical systems. Recently, attempts to deal with some of these issues have been made, \eg, the MOT Benchmark \cite{MOT16} provides a large set of annotated testing video sequences, unified detection hypotheses, standard evaluation tools, \etc This is very likely to advance the further studies and developments of MOT techniques.

\subsection{Future Directions}
\label{ssec_directions}
Even after decades of research on the MOT problem, there are still numerous research opportunities in studying this problem. Here we would like to point out some more prevalent problems and provide possible research directions.

\textbf{\emph{MOT with video adaptation.}} As mentioned above, the majority of current MOT methods requires an offline trained object detector. There arises a problem that the detection result for a specific video is not optimal since the object detector is not trained for the given video. This often limits the performance of multiple object tracking. A customization of the object detector is necessary to improve MOT performance. One solution proposed by Shu \etal~\cite{shu2013improving} adapts a generic pedestrian detector to a specific video by progressively refining the generic pedestrian detector. This is one important direction to follow in order to improve the pre-processing stage for MOT methods.
	
\textbf{\emph{MOT under multiple cameras.}} It is obvious that MOT would benefit from multi-camera settings~\cite{ECCVW16MOT,Ristani_2018_CVPR}. There are two kinds of configurations of multiple cameras. The first one is that multiple cameras record the same scene, \ie, multiple views. However, one key question in this setting is how to fuse the information from multiple cameras. The second one is that each camera records a different scene, \ie, a non-overlapping multi-camera network. In that case, the data association across multiple cameras becomes a re-identification problem.
	
\textbf{\emph{Multiple 3D object tracking.}} Most of the current approaches focus on multiple object tracking in 2D, \ie, on the image plane, even in the case of multiple cameras. 3D tracking~\cite{park2008multiple}, which could provide more accurate position, size estimation and effective occlusion handling for high-level computer vision tasks, is potentially more useful. However, 3D tracking requires camera calibration, or has to overcome other challenges for estimating camera poses and scene layout. Meanwhile, 3D model design is another issue exclusive to 2D MOT.

\textbf{\emph{MOT with scene understanding.}} Previous studies ~\cite{rodriguez2011data, zhou2012understanding, zhou2012coherent} have been performed to analyze over-crowded scenarios such as underground train stations during peak hours and demonstrations in public places. In this kind of scenarios, most objects are small and/or largely occluded, thus are very difficult to track. The analyzing results from scene understanding can provide contextual information and scene structure, which is very helpful to the tracking problem if it is better incorporated into an MOT algorithm.
	
\textbf{\emph{MOT with deep learning.}} Deep learning based models have emerged as an extremely powerful framework to deal with different kinds of vision problems including image classification \cite{krizhevsky2012imagenet}, object detection \cite{NIPS15FasterRCNN,CVPR16YOLO,ECCV16SSD}, and more relevantly single object tracking \cite{CVPR14RCNN}. For the MOT problem, the strong observation model provided by the deep learning model for target detection can boost the tracking performance significantly \cite{ECCVW16POI,ECCVW16MCMOT}. The formulation and modeling of the target association problem using deep neural networks \cite{Son_2017_CVPR,Schulter_2017_CVPR,Sadeghian_2017_ICCV,Chu_2017_ICCV} need more research efforts, although the first attempt to employ sequential neural networks for online MOT has been made very recently. The modules like attention mechanism \cite{Zhu_2018_ECCV}, LSTM \cite{Maksai_2019_CVPR,Kim_2018_ECCV} are also employed by researchers for solving the MOT problem.
	
\textbf{\emph{MOT with other computer vision tasks.}} Although multiple object tracking serves other high-level computer vision tasks, there is a trend to solve multi-object tracking with some other computer vision tasks jointly as they are beneficial to each other. Possible combinations include object segmentation \cite{CVPR15TrackFCN,Milan_2015_CVPR,Koh_2017_ICCV,Voigtlaender_2019_CVPR}, re-identification \cite{ECCV16DLAttributeReID,Ristani_2018_CVPR,Tang_2019_CVPR}, human pose estimation \cite{ICCV15PoseVideo,Insafutdinov_2017_CVPR,PoseTrack,Jin_2019_CVPR,Raaj_2019_CVPR}, and action recognition \cite{choi2012unified}.

Besides the above future directions, since the current MOT research is mainly focused on tracking multiple humans in a surveillance scenario, the extensions of the current MOT research to other types of targets (\eg, vehicles, animals, \etc) and scenarios (\eg, traffic scenes, aerial photographs, \etc) are also very good research directions, since the problem settings and difficulties for tracking different types of targets under different scenarios are sometimes quite different from those in tracking multiple humans in a surveillance scenario.

\setlength{\bibsep}{0.5ex}
\bibliography{motsurvey}

\end{document}